\documentclass[journal]{assets/IEEEtran}

\usepackage{cite}
\usepackage{array} 
\newcolumntype{P}[1]{>{\centering\arraybackslash}p{#1}} 
\usepackage{mathtools} 
\usepackage[flushleft]{threeparttable} 
\usepackage{amsmath,amssymb,amsfonts}
\usepackage{graphicx}
\usepackage{textcomp}
\makeatletter
\let\MYcaption\@makecaption
\makeatother
\usepackage[font=footnotesize]{subcaption}
\makeatletter
\let\@makecaption\MYcaption
\makeatother
\usepackage{enumitem}
\usepackage{xcolor}
\usepackage{url} 
\usepackage{tikz} 
\usepackage{pgfplots} 
\usepackage[final,poster=last]{animate} 
\usepackage{multirow}
\usepackage{color,soul} 
\soulregister\cite7 
\soulregister\ref7 

\definecolor{small_2d_cnn_flow}{RGB}{255,222,0}
\definecolor{small_2d_cnn_frame}{RGB}{0,255,19}
\definecolor{small_3d_cnn}{RGB}{69,91,255}
\definecolor{small_cnn_lstm}{RGB}{255,149,0}
\definecolor{small_two_stream}{RGB}{255,0,230}
\definecolor{small_slowfast}{RGB}{0,245,255}
\definecolor{resnet_2d_cnn_flow}{RGB}{178,155,0}
\definecolor{resnet_2d_cnn_frame}{RGB}{0,155,12}
\definecolor{resnet_3d_cnn}{RGB}{0,18,155}
\definecolor{resnet_cnn_lstm}{RGB}{153,89,0}
\definecolor{resnet_two_stream}{RGB}{153,0,138}
\definecolor{resnet_slowfast}{RGB}{0,147,153}

\begin{document}

\title{Learning deep representations for video-based intake gesture detection}
\author{Philipp~V.~Rouast,~\IEEEmembership{Student~Member,~IEEE,}
        Marc~T.~P.~Adam%
\thanks{The authors are with the School of Electrical Engineering and Computing, The University of Newcastle, Callaghan, NSW 2308, Australia. E-mail: philipp.rouast@uon.edu.au, marc.adam@newcastle.edu.au.}%
}

\markboth{Journal of \LaTeX\ Class Files,~Vol.~14, No.~8, August~2015}%
{Shell \MakeLowercase{\textit{et al.}}: Bare Demo of IEEEtran.cls for IEEE Journals}

\maketitle

\begin{abstract}
Automatic detection of individual intake gestures during eating occasions has the potential to improve dietary monitoring and support dietary recommendations.
Existing studies typically make use of on-body solutions such as inertial and audio sensors, while video is used as ground truth.
Intake gesture detection directly based on video has rarely been attempted.
In this study, we address this gap and show that deep learning architectures can successfully be applied to the problem of video-based detection of intake gestures.
For this purpose, we collect and label video data of eating occasions using 360-degree video of 102 participants.
Applying state-of-the-art approaches from video action recognition, our results show that (1) the best model achieves an $F_1$ score of 0.858, (2) appearance features contribute more than motion features, and (3) temporal context in form of multiple video frames is essential for top model performance.
\end{abstract}

\begin{IEEEkeywords}
Deep learning, intake gesture detection, dietary monitoring, video-based
\end{IEEEkeywords}

\section{Introduction}
\label{sec:introduction}

\IEEEPARstart{D}{ietary} monitoring plays an important role in assessing an individual's overall dietary intake and, based on this, providing targeted dietary recommendations.
Dietitians \cite{weekes2009review} and personal monitoring solutions \cite{rouast2018using} rely on accurate dietary information to support individuals in meeting their health goals.
For instance, research has shown that the global risk and burden of non-communicable disease is associated with poor diet and hence requires targeted interventions \cite{who2017non}.
However, manually assessing dietary intake often involves considerable processing time and is subject to human error \cite{lichtman1992discrepancy}.

Automatic dietary monitoring aims to detect (i) \textit{when}, (ii) \textit{what}, and (iii) \textit{how much} is consumed \cite{vu2017wearable}.
This is a complex and multi-faceted problem involving tasks such as action detection to identify intake gestures (\textit{when}), object recognition and segmentation to identify individual foods (\textit{what}), as well as volume and density estimation to derive food quantity (\textit{how much}).
A variety of sensors have been explored in the literature, including inertial, audio, visual, and piezoelectric sensors \cite{vu2017wearable}, \cite{kyritsis2019modeling}, \cite{hantke2016hear}.

Detection of individual intake gestures can improve detection of entire eating occasions \cite{thomaz2015practical} and amounts consumed \cite{mirtchouk2016automated}.
It also provides access to measures such as intake speed, as well as meta-information for easier review of videos.
Although video is often used as ground truth for studies focused on detecting chews, swallows, and intake gestures, it has rarely been used as the basis for automatic detection.
However, there are several indications that video could be a suitable data source to monitor such events:
(i) increasing exploration of video monitoring in residential and hospital settings \cite{braeken2016secure}, \cite{hall2017implementing},
(ii) the rich amount of information embedded in the visual modality, and
(iii) recent advances in machine learning, and in particular deep learning \cite{lecun2015deep}, for video action recognition that have largely been left unexplored in dietary monitoring.

In this paper, we address this gap by demonstrating the feasibility of using deep neural networks (DNNs) for automatic detection of intake gestures from raw video frames.
For this purpose, we investigate the 3D CNN \cite{ji20133d}, CNN-LSTM \cite{donahue2015long}, Two-Stream \cite{simonyan2014two}, and SlowFast \cite{feichtenhofer2018slowfast} architectures which have been applied in the field of video action recognition, but not for dietary monitoring.
These architectures allow to consider \textit{temporal context} in the form of multiple frames.
Further, instead of relying on handcrafted models and features, deep learning leverages a large number of examples to learn feature representations on multiple levels of abstraction.
In dietary monitoring, deep learning has mainly been used for image-based food recognition (\textit{what}) \cite{ciocca2017learning}, and recently in intake gesture detection based on inertial sensors (\textit{when}) \cite{kyritsis2019modeling}.
However, it has yet to be applied on video-based intake gesture detection.
Our main contributions are the following:

\begin{enumerate}
	\item We fill the gap between dietary monitoring and video action recognition by demonstrating the feasibility of using deep learning architectures to detect individual intake gestures from raw video frames. We conduct a laboratory study with 102 participants and 4891 intake gestures, by sourcing video from a 360-degree camera placed in the center of the table. A ResNet-50 SlowFast model achieved the best $F_1$ score of 0.858.
	\item Video action recognition can build on both appearance and motion features. It is in general not clear which are more important for a given action \cite{feichtenhofer2018have}. Using a 2D CNN without temporal context, we show that appearance (individual frames) performs better than motion (optical flow between adjacent frames) for detecting intake gestures.
	\item Similarly, it is not clear to what extent temporal context improves model accuracy of detecting a given action \cite{karpathy2014large}. Comparing the best model \textit{with} (ResNet-50 SlowFast) and the best 2D CNN \textit{without} temporal context, we find a relative $F_1$ improvement of 8\%.

\end{enumerate}

The remainder of the paper is organized as follows:
In Section \ref{sec:related}, we discuss the related literature, including dietary monitoring and video action recognition.
Our proposed models are introduced in Section \ref{sec:method}, and the dataset in Section \ref{sec:dataset}.
We present our experiments and results in Section \ref{sec:experiments}, and draw conclusions in Section \ref{sec:conclusion}.


\begin{figure*}[t]
\centering
\includegraphics[width=.85\textwidth]{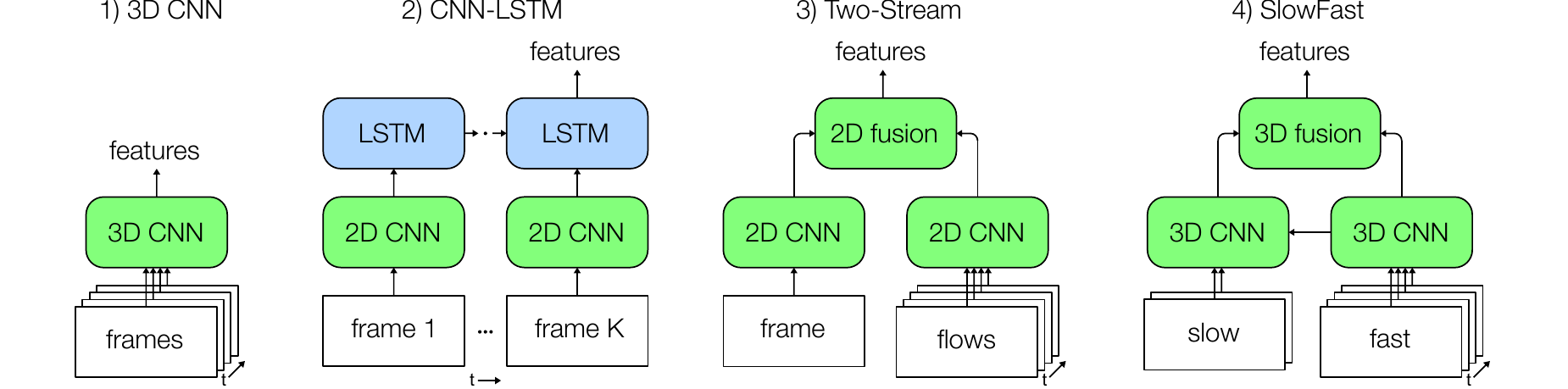}
\caption{The four investigated approaches from video action recognition based on temporal context $t$ (adapted from Carreira and Zisserman \cite{carreira2017quo}).}
\label{fig:architectures}
\end{figure*}

\section{Related Research}
\label{sec:related}

\subsection{Dietary monitoring}
\label{sec:related:sub:food-intake}

At the conceptual level, dietary monitoring broadly captures three components of recognition, namely \textit{what} (e.g., identification of specific foods), \textit{how much} (i.e., quantification of consumed food), and \textit{when} (i.e., timing of eating occasions).
Traditional paper-based methods such as recalls and specialized questionnaires \cite{block1982review} are still commonly used by dietitians.
Amongst end-users, mobile applications that allow manual logging of individual meals are also popular.
These active methods are characterized by a considerable amount of effort, and known to be affected by biases and human error \cite{lichtman1992discrepancy}.

Realizing the requirement for objective measurements of a person's diet, several sensor-based approaches of passively collecting information associated with diet have been proposed in the literature.
With the emergence of labeled databases of food images \cite{chen2016deep}, \cite{ciocca2017learning}, food recognition from still images has become a popular task in computer vision research.
The state of the art uses features learned by deep convolutional neural networks (CNNs) to distinguish between food classes \cite{ciocca2018cnn}.
CNNs are DNNs especially designed for visual inputs.

Image-based estimation of food volume and associated calories typically extends food recognition by volume estimation of different foods, and linking with nutrient databases \cite{puri2009recognition}, \cite{zhang2015snap}.
Estimation of food volume from audio and inertial sensors based on individual bite sizes has also been proposed \cite{mirtchouk2016automated}.

In detecting intake behavior, we distinguish between detection of events describing meal microstructure (e.g., individual intake gestures), and detecting intake occasions as a whole (e.g., a meal), which can be seen as clusters of detected events \cite{dong2014detecting}.
Besides aiding in the estimation of food volume \cite{mirtchouk2016automated}, information about meal microstructure can be leveraged to improve active methods \cite{ye2016assisting}.
It also allows dietitians to quantify measures of interest such as the rate of eating \cite{robinson2014systematic}.

In general, detection of chews and swallows is typically attempted using on-body audio or piezoelectric sensors, whilst detection of intake gestures is the domain of wrist-mounted inertial sensors \cite{heydarian2019assessing}.
Chews and swallows generate characteristic audio signatures, which was exploited for automatic detection of meal microstructure as early as 2005 \cite{amft2005analysis}, \cite{passler2012food}.
Swallows can also be registered using piezoelectric sensors measuring strain on the jaw \cite{sazonov2012sensor}.
Inertial sensors can be used to measure the acceleration and spatial movements of the wrist to identify intake gestures \cite{amft2005detection}, \cite{shen2017assessing}, \cite{zhang2018sense}.
Recently, DNNs were applied for this purpose \cite{kyritsis2019modeling}.

\subsection{Video-based intake gesture recognition}
\label{sec:related:sub:video-based-gesture-recognition}

Despite the importance of visual sensors for recording ground truth, video data of eating occasions is rarely considered as the basis for automatic detection of meal microstructure.
This is surprising, as the visual modality contains a broad range of information about intake behavior.
In fact, in 2004, one of the earliest works in this field considered surveillance type video recorded in a nursing home to detect intake gestures \cite{gao2004dining}.
This approach relied on optical flow-based motion features, which were used to train a Hidden Markov Model.
A further approach used object detection of face, mouth, and eating utensils which was realised with haar-like appearance features \cite{okamoto2016grillcam}.
We also see skeleton-based approaches with additional depth information \cite{hondori2012monitoring}, \cite{tham2014automatic}.
Deep learning, which is the state of the art for video action recognition, has not been explored to the best of our knowledge.

\subsection{Video action recognition}
\label{sec:related:sub:video-action-recognition}

The task of action recognition from video extends 2D image input by the dimension of time.
While temporal context can carry useful information, it also complicates the search for good feature representations given the typically much larger dimensionality of the raw data.
Before the proliferation of deep learning, approaches in video action recognition would follow the traditional paradigm of pattern recognition:
Computing complex hand-crafted features from raw video frames, based on which shallow classifiers could be learned.
Such features were either video-level aggregation of local spatio-temporal features such as HOG3D \cite{klaser2008spatio}, or point trajectories of dense points computed, e.g., using optical flow \cite{wang2011action}.
The following four deep learning architectures emerged from the literature on video action recognition, as shown in Fig. \ref{fig:architectures}:

\subsubsection{3D CNN -- Spatio-temporal convolutions}

The 3D CNN approach features 3D convolutions instead of the 2D convolutions found in 2D CNNs.
Videos are treated as spatio-temporal volumes, where the third dimension represents temporal context.
3D CNNs can thus automatically learn low-level features that take into account both spatial and temporal information.
This approach was first proposed in 2010 by Ji et al. \cite{ji20133d}, who integrated 3D convolutions with handcrafted features.
Running experiments with end-to-end training on larger datasets, Karpathy et al. \cite{karpathy2014large} reported that it works best to slowly fuse the temporal information throughout the network.
However, they found that temporal context only improves model accuracy for some classes such as juggling; furthermore, it reduced accuracy for some classes \cite{karpathy2014large}.
Other experiments regarding architecture choices concluded that 3D CNN can model appearance and motion simultaneously \cite{tran2015learning}.

\subsubsection{CNN-LSTM -- Incorporating recurrent neural networks}

In the CNN-LSTM approach, the temporal context is modelled by a recurrent neural network (RNN).
RNNs are DNNs that take the previous model state as an additional input.
In 2015, Donahue et al. \cite{donahue2015long} proposed to use the sequence of high-level spatial features learned by a CNN from individual video frames as input into a long short-term memory (LSTM) RNN.
Such LSTM networks are known to be easier to train for longer sequences \cite{hochreiter1997long}.
The CNN-LSTM model has the advantage of being more flexible with regards to the number of input frames, but has relatively many parameters and appears to be more data hungry in comparison to other approaches \cite{carreira2017quo}.

\subsubsection{Two-Stream -- Decoupling appearance and motion}

In 2014, Simonyan and Zisserman \cite{simonyan2014two} observed that 2D CNN models without temporal context achieved accuracy close to the 3D CNN approach \cite{karpathy2014large}, and that state-of-the-art accuracies involved handcrafted trajectory-based representations based on motion features.
They proposed the two-stream architecture, which decouples appearance and motion by using a single still frame (appearance) and temporal context in form of stacked optical flow (motion).
Both are fed into separate CNNs, where the appearance CNN is pre-trained on the large ImageNet database.
While the original design employed score-level fusion \cite{simonyan2014two}, later variants used feature-level fusion of the last CNN layers \cite{feichtenhofer2016convolutional}.

\subsubsection{SlowFast -- Joint learning at different temporal resolutions}

The SlowFast architecture proposed by Feichtenhofer et al. \cite{feichtenhofer2018slowfast} in late 2018 learns from temporal context at multiple temporal resolutions.
As of mid 2019, it represents the state of the art in video action recognition with 79\% accuracy on the large Kinetics dataset without any pre-training.
The idea of decoupling slow and fast motion is integrated into the network design.
Two pathways make up the SlowFast architecture, consisting of a 3D CNN each: The \textit{slow} pathway has more capacity to learn about appearance than motion, while the \textit{fast} pathway works the other way around.
This is realized by setting a factor $\alpha$ as the difference in sequence downsampling, and a factor $\beta$ as the difference in learned channels.
A number of lateral connections allow that information from both pathways is fused.


\begin{figure}[t]
\centering
\includegraphics[width=.95\columnwidth]{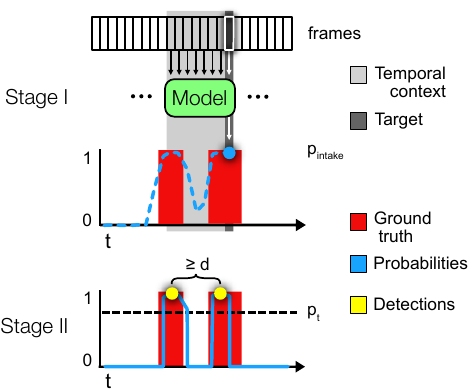}
\caption{Illustration of sample outputs at the two stages. In Stage I, the model estimates the probability $p_{intake}$ for a target frame. For models with temporal context, input consists of multiple frames (16 in our experiments), of which the last frame is the target. In Stage II, detections of intake events are realized using a local maximum search on the $p_t$-thresholded series of probabilities, where the detections have to be at least $d$ apart (in our experiments, $d = 2s$).}
\label{fig:stages}
\end{figure}

\section{Proposed Method}
\label{sec:method}

Detecting individual intake gestures from video requires prediction of sparse points in time.
We adopt the approach of Kyritsis et al. \cite{kyritsis2019modeling} and split this problem into two stages, as illustrated in Fig. \ref{fig:stages}:

\begin{description}
	\item[Stage I:] \textbf{Estimation of state probability at the frame level}, i.e., estimating the probability $p_{intake}$ for each frame, and
	\item[Stage II:] \textbf{Detection of intake gestures}, by selecting sparse points in time based on the estimated probabilities.
\end{description}

\makeatletter
\newcommand{\thickhline}{%
    \noalign {\ifnum 0=`}\fi \hrule height 1pt
    \futurelet \reserved@a \@xhline
}
\newcolumntype{"}{@{\hskip\tabcolsep\vrule width 1pt\hskip\tabcolsep}}
\makeatother
\newcommand{\slow}[1]{\textcolor{orange}{#1}}
\newcommand{\fast}[1]{\textcolor{cyan}{#1}}
\newcommand{\frme}[1]{\textcolor{red}{#1}}
\newcommand{\flow}[1]{\textcolor{blue}{#1}}

\begin{table*}[t]
\caption{\textbf{Small model instantiations}. We report temporal, spatial, and channel dim. of convolution kernels as $\{T\times S^2$, $C\}$, temporal and spatial dim. of pooling ops. as $\{T \times S^2\}$, and stride sizes likewise. Corresponding output sizes are reported as $\{T\times S^2\times C\}$.}
\label{tab:small}
\setlength{\extrarowheight}{2pt}
\begin{threeparttable}
\begin{tabular}{ p{.7cm} " p{.6cm} | p{1.2cm} " p{1.1cm} | p{1.6cm} " p{.6cm} | p{1.8cm} " p{.6cm} | p{1.2cm} " p{1.7cm} | p{1.6cm} }
& \multicolumn{2}{c"}{0a) 2D CNN\tnote{a}} & \multicolumn{2}{c"}{1a) 3D CNN} & \multicolumn{2}{c"}{2a) CNN-LSTM} & \multicolumn{2}{c"}{3a) Two-Stream\tnote{a}} & \multicolumn{2}{c}{4a) SlowFast\tnote{b}} \\
Layer & \multicolumn{1}{c}{} & \parbox[t]{1cm}{\frme{frame}\\\flow{flow}} & \multicolumn{1}{c}{} & & \multicolumn{1}{c}{} & & \multicolumn{1}{c}{} & \parbox[t]{1.3cm}{\frme{frame}\\\flow{flows}} & \multicolumn{1}{c}{} & \parbox[t]{1.8cm}{\slow{slow}\\\fast{fast}} \\ [.3cm]
\thickhline
data & & \parbox[t]{1.3cm}{$128^2\times\frme{3}|\flow{2}$} & & $16\times128^2\times3$ & & $16\times128^2\times3$ & & \parbox[t]{1.3cm}{\frme{$128^2\times3$}\\\flow{$128^2\times32$}} & & \parbox[t]{1.8cm}{\slow{$4\times128^2\times3$}\\\fast{$16\times128^2\times3$}} \\ [.3cm]
\hline
conv0\tnote{c} & & & & & & & \parbox[t]{1cm}{$3^2,3$\\str. $1^2$} & \parbox[t]{1.3cm}{\frme{$128^2\times3$}\\\flow{$128^2\times3$}} & & \\ [.3cm]
\hline
conv1 & \parbox[t]{.8cm}{$3^2,32$\\str. $1^2$} & \parbox[t]{1.2cm}{$128^2\times32$} & \parbox[t]{1.2cm}{$3\times3^2,32$\\str. $1\times1^2$} & \parbox[t]{1.7cm}{$16\times128^2\times32$} & \parbox[t]{1cm}{$3^2,32$\\str. $1^2$} & \parbox[t]{2cm}{$16\times128^2\times32$} & \parbox[t]{1cm}{$3^2,32$\\str. $1^2$} & \parbox[t]{1.3cm}{\frme{$128^2\times32$}\\\flow{$128^2\times32$}} & \parbox[t]{2cm}{$\slow{1}|\fast{3}\times3^2,\slow{32}|\fast{8}$\\str. $1\times1^2$} & \parbox[t]{1.8cm}{\slow{$4\times128^2\times32$}\\\fast{$16\times128^2\times8$}} \\ [.3cm]
\hline
pool1 & \parbox[t]{.7cm}{$2^2$\\str. $2^2$} & $64^2\times32$ & \parbox[t]{1.1cm}{$2\times2^2$\\str. $2\times2^2$} & $8\times64^2\times32$ & \parbox[t]{.7cm}{$2^2$\\str. $2^2$} & $16\times64^2\times32$ & \parbox[t]{.7cm}{$2^2$\\str. $2^2$} & \parbox[t]{1.5cm}{\frme{$64^2\times32$}\\\flow{$64^2\times32$}} & \parbox[t]{1.5cm}{$1\times2^2$\\str. $1\times2^2$} & \parbox[t]{1.8cm}{\slow{$4\times64^2\times32$}\\\fast{$16\times64^2\times8$}} \\ [.3cm]
\hline
conv2 & \parbox[t]{.8cm}{$3^2,32$\\str. $1^2$} & $64^2\times32$ & \parbox[t]{1.2cm}{$3\times3^2,32$\\str. $1\times1^2$} & $8\times64^2\times32$ & \parbox[t]{1cm}{$3^2,32$\\str. $1^2$} & $16\times64^2\times32$ & \parbox[t]{1cm}{$3^2,32$\\str. $1^2$} & \parbox[t]{1.5cm}{\frme{$64^2\times32$}\\\flow{$64^2\times32$}} & \parbox[t]{2cm}{$\slow{1}|\fast{3}\times3^2,\slow{32}|\fast{8}$\\str. $1\times1^2$} & \parbox[t]{1.8cm}{\slow{$4\times64^2\times32$}\\\fast{$16\times64^2\times8$}} \\ [.3cm]
\hline
pool2 & \parbox[t]{.7cm}{$2^2$\\str. $2^2$} & $32^2\times32$ & \parbox[t]{1.1cm}{$2\times2^2$\\str. $2\times2^2$} & $4\times32^2\times32$ & \parbox[t]{.7cm}{$2^2$\\str. $2^2$} & $16\times32^2\times32$ & \parbox[t]{.7cm}{$2^2$\\str. $2^2$} & \parbox[t]{1.5cm}{\frme{$32^2\times32$}\\\flow{$32^2\times32$}} & \parbox[t]{1.5cm}{$1\times2^2$\\str. $1\times2^2$} & \parbox[t]{1.8cm}{\slow{$4\times32^2\times32$}\\\fast{$16\times32^2\times8$}} \\ [.3cm]
\hline
conv3 & \parbox[t]{.8cm}{$3^2,64$\\str. $1^2$} & $32^2\times64$ & \parbox[t]{1.2cm}{$3\times3^2,64$\\str. $1\times1^2$} & $4\times32^2\times64$ & \parbox[t]{1cm}{$3^2,64$\\str. $1^2$} & $16\times32^2\times64$ & \parbox[t]{1cm}{$3^2,64$\\str. $1^2$} & \parbox[t]{1.5cm}{\frme{$32^2\times64$}\\\flow{$32^2\times64$}} & \parbox[t]{2cm}{$\slow{1}|\fast{3}\times3^2,\slow{64}|\fast{16}$\\str. $1\times1^2$} & \parbox[t]{1.8cm}{\slow{$4\times32^2\times64$}\\\fast{$16\times32^2\times16$}} \\ [.3cm]
\hline
pool3 & \parbox[t]{.7cm}{$2^2$\\str. $2^2$} & $16^2\times64$ & \parbox[t]{1.1cm}{$2\times2^2$\\str. $2\times2^2$} & $2\times16^2\times64$ & \parbox[t]{.7cm}{$2^2$\\str. $2^2$} & $16\times16^2\times64$ & \parbox[t]{.7cm}{$2^2$\\str. $2^2$} & \parbox[t]{1.5cm}{\frme{$16^2\times64$}\\\flow{$16^2\times64$}} & \parbox[t]{1.5cm}{$1\times2^2$\\str. $1\times2^2$} & \parbox[t]{1.8cm}{\slow{$4\times16^2\times64$}\\\fast{$16\times16^2\times16$}} \\ [.3cm]
\hline
conv4 & \parbox[t]{.8cm}{$3^2,64$\\str. $1^2$} & $16^2\times64$ & \parbox[t]{1.2cm}{$3\times3^2,64$\\str. $1\times1^2$} & $2\times16^2\times64$ & \parbox[t]{1cm}{$3^2,64$\\str. $1^2$} & $16\times16^2\times64$ & \parbox[t]{1cm}{$3^2,64$\\str. $1^2$} & \parbox[t]{1.5cm}{\frme{$16^2\times64$}\\\flow{$16^2\times64$}} & \parbox[t]{2cm}{$\slow{1}|\fast{3}\times3^2,\slow{64}|\fast{16}$\\str. $1\times1^2$} & \parbox[t]{1.8cm}{\slow{$4\times16^2\times64$}\\\fast{$16\times16^2\times16$}} \\ [.3cm]
\hline
pool4 & \parbox[t]{.7cm}{$2^2$\\str. $2^2$} & $8^2\times64$ & \parbox[t]{1.1cm}{$2\times2^2$\\str. $2\times2^2$} & $1\times8^2\times64$ & \parbox[t]{.7cm}{$2^2$\\str. $2^2$} & $16\times8^2\times64$ & \parbox[t]{.7cm}{$2^2$\\str. $2^2$} & \parbox[t]{1.5cm}{\frme{$8^2\times64$}\\\flow{$8^2\times64$}} & \parbox[t]{1.5cm}{$1\times2^2$\\str. $1\times2^2$} & \parbox[t]{1.8cm}{\slow{$4\times8^2\times64$}\\\fast{$16\times8^2\times16$}} \\ [.3cm]
\hline
fusion & & & & & & & & $8^2\times64$ & & $8^2\times64$ \\
\hline
flatten & & $4096$ & & $4096$ & & $16\times4096$ & & $4096$ & & $4096$ \\
\hline
dense & & $1024$ & & $1024$ & & $16\times1024$ & & $1024$ & & $1024$ \\
\hline
lstm & & & & & & $16\times128$ & & & & \\
\hline
dense & & $2$ & & $2$ & & $16\times 2$ & & $2$ & & $2$ \\
\hline
\end{tabular}
\begin{tablenotes}
\item[a] For 2D CNN and Two-Stream, colors \frme{red} and \flow{blue} highlight how dimensions differ between \frme{frames} and \flow{flows}.
\item[b] For SlowFast, colors $\slow{\text{orange}}|\fast{\text{cyan}}$ highlight the differences in model parameters and dimensions between the \slow{slow} and \fast{fast} pathways.
\item[c] Only for flow input; Serves the purpose of producing 3 channels for transfer learning.
\end{tablenotes}
\end{threeparttable}
\end{table*}

\begin{table*}[t]
\caption{\textbf{ResNet-50 model instantiations}. We report temporal, spatial, and channel dim. of conv. kernels as $\{T\times S^2$, $C\}$, temporal and spatial dim. of pooling ops. as $\{T \times S^2\}$, and stride sizes likewise. The corresponding output sizes are reported as $\{T\times S^2\times C\}$.}
\label{tab:resnet}
\setlength{\extrarowheight}{3pt}
\newcommand{\ns}{\negthickspace}
\begin{threeparttable}
\begin{tabular}{ p{.7cm} " p{.9cm} | p{.7cm} " p{1.4cm} | p{1.1cm} " p{.9cm} | p{1.2cm} " p{.9cm} | p{1.3cm} " p{1.9cm} | p{1.8cm} }
Layer & \multicolumn{2}{c"}{0b) 2D CNN\tnote{a}} & \multicolumn{2}{c"}{1b) 3D CNN} & \multicolumn{2}{c"}{2b) CNN-LSTM} & \multicolumn{2}{c"}{3b) Two-Stream\tnote{a}} & \multicolumn{2}{c}{4b) SlowFast\tnote{b}} \\
& \multicolumn{1}{c}{} & \parbox[t]{.7cm}{\frme{frame}\\\flow{flow}} & \multicolumn{1}{c}{} & & \multicolumn{1}{c}{} & & \multicolumn{1}{c}{} & \parbox[t]{1cm}{\frme{frame}\\\flow{flows}} & \multicolumn{1}{c}{} & \parbox[t]{1cm}{\slow{slow}\\\fast{fast}} \\ [.3cm]
\thickhline
data & & \parbox[t]{.6cm}{$224^2$\\$\times\frme{3}|\flow{2}$} & & \parbox[t]{1.2cm}{$16\times128^2$\\$\times3$} & & \parbox[t]{1.2cm}{$16\times224^2$\\$\times3$} & & \parbox[t]{1.5cm}{\frme{$224^2\times3$}\\\flow{$224^2\times32$}} & & \parbox[t]{2cm}{\slow{$2\times128^2\times3$}\\\fast{$16\times128^2\times3$}} \\ [.3cm]
\hline
conv0\tnote{c} & \parbox[t]{1cm}{$3^2, 3$\\stride $1^2$} & \parbox[t]{.7cm}{$112^2$\\$\times3$} & & & & & \parbox[t]{1cm}{$3^2,3$\\stride $1^2$} & \parbox[t]{1.5cm}{\frme{$224^2\times3$}\\\flow{$224^2\times3$}} & & \\ [.3cm]
\hline
conv1 & \parbox[t]{1cm}{$7^2,64$\\stride $2^2$} & \parbox[t]{.6cm}{$112^2$\\$\times64$} & \parbox[t]{1.5cm}{$3\times5^2,64$\\stride $1\times1^2$} & \parbox[t]{1.2cm}{$16\times128^2$\\$\times64$} & \parbox[t]{1cm}{$7^2,64$\\stride $2^2$} & \parbox[t]{1.2cm}{$16\times112^2$\\$\times64$} & \parbox[t]{1cm}{$7^2,64$\\stride $2^2$} & \parbox[t]{1.5cm}{\frme{$112^2\times64$}\\\flow{$112^2\times64$}} & \parbox[t]{1.8cm}{$\slow{1}|\fast{3}\times5^2,\slow{64}|\fast{8}$\\stride $1\times1^2$} & \parbox[t]{1.8cm}{\slow{$2\times128^2\times64$}\\\fast{$16\times128^2\times8$}} \\ [.3cm]
\hline
pool1 & \parbox[t]{1cm}{$3^2$\\stride $2^2$} & \parbox[t]{.6cm}{$56^2$\\$\times64$} & \parbox[t]{1.5cm}{$3\times3^2$\\stride $2\times2^2$} & \parbox[t]{1.2cm}{$8\times64^2$\\$\times64$} & \parbox[t]{1cm}{$3^2$\\stride $2^2$} & \parbox[t]{1.2cm}{$16\times56^2$\\$\times64$} & \parbox[t]{1cm}{$3^2$\\stride $2^2$} & \parbox[t]{1.5cm}{\frme{$56^2\times64$}\\\flow{$56^2\times64$}} & \parbox[t]{1.8cm}{$1\times3^2$\\stride $1\times2^2$} & \parbox[t]{1.8cm}{\slow{$2\times64^2\times64$}\\\fast{$16\times64^2\times8$}} \\ [.3cm]
\hline
\parbox[t]{.4cm}{res2.x\\($\times3$)} & $\begin{matrix} 1^2, 64 \\ 3^2, 64 \\ 1^2, 256 \end{matrix}$ & \parbox[t]{.6cm}{$56^2$\\$\times256$} & $\begin{matrix} 3\times1^2, 64 \\ 1\times3^2, 64 \\ 1\times1^2, 256 \end{matrix}$ & \parbox[t]{1.2cm}{$8\times64^2$\\$\times256$} & $\begin{matrix} 1^2, 64 \\ 3^2, 64 \\ 1^2, 256 \end{matrix}$ & \parbox[t]{1.2cm}{$16\times56^2$\\$\times256$} & $\begin{matrix} 1^2, 64 \\ 3^2, 64 \\ 1^2, 256 \end{matrix}$ & \parbox[t]{1.5cm}{\frme{$56^2\times256$}\\\flow{$56^2\times256$}} & $\begin{matrix} \slow{1}|\fast{3}\times1^2, \slow{64}|\fast{8} \\ 1\times3^2, \slow{64}|\fast{8} \\ 1\times1^2, \slow{256}|\fast{32} \end{matrix}$ & \parbox[t]{1.8cm}{\slow{$2\times64^2\times256$}\\\fast{$16\times64^2\times32$}} \\ [.5cm]
\hline
\parbox[t]{.4cm}{res3.x\\($\times4$)} & $\begin{matrix} 1^2, 128 \\ 3^2, 128 \\ 1^2, 512 \end{matrix}$ & \parbox[t]{.6cm}{$28^2$\\$\times512$} & $\begin{matrix} 3\times1^2, 128 \\ 1\times3^2, 128 \\ 1\times1^2, 512 \end{matrix}$ & \parbox[t]{1.2cm}{$4\times32^2$\\$\times512$} & $\begin{matrix} 1^2, 128 \\ 3^2, 128 \\ 1^2, 512 \end{matrix}$ & \parbox[t]{1.2cm}{$16\times28^2$\\$\times512$} & $\begin{matrix} 1^2, 128 \\ 3^2, 128 \\ 1^2, 512 \end{matrix}$ & \parbox[t]{1.5cm}{\frme{$28^2\times512$}\\\flow{$28^2\times512$}} & $\begin{matrix} \slow{1}|\fast{3}\times1^2, \slow{128}|\fast{16} \\ 1\times3^2, \slow{128}|\fast{16} \\ 1\times1^2, \slow{512}|\fast{64} \end{matrix}$ & \parbox[t]{1.8cm}{\slow{$2\times32^2\times512$}\\\fast{$16\times32^2\times64$}} \\ [.5cm]
\hline
\parbox[t]{.4cm}{res4.x\\($\times6$)} & $\ns\begin{matrix} 1^2, 256 \\ 3^2, 256 \\ 1^2, 1024 \end{matrix}$ & \parbox[t]{.6cm}{$14^2$\\$\times1024$} & $\ns\begin{matrix} 3\times1^2, 256 \\ 1\times3^2, 256 \\ 1\times1^2, 1024 \end{matrix}$ & \parbox[t]{1.2cm}{$2\times16^2$\\$\times1024$} & $\ns\begin{matrix} 1^2, 256 \\ 3^2, 256 \\ 1^2, 1024 \end{matrix}$ & \parbox[t]{1.2cm}{$16\times14^2$\\$\times1024$} & $\ns\begin{matrix} 1^2, 256 \\ 3^2, 256 \\ 1^2, 1024 \end{matrix}$ & \parbox[t]{1.5cm}{\frme{$14^2\times1024$}\\\flow{$14^2\times1024$}} & $\ns\begin{matrix} 3\times1^2, \slow{256}|\fast{32} \\ 1\times3^2, \slow{256}|\fast{32} \\ 1\times1^2, \slow{1024}|\fast{128} \end{matrix}$ & \parbox[t]{1.8cm}{\slow{$2\times16^2\times1024$}\\\fast{$16\times16^2\times128$}} \\ [.5cm]
\hline
\parbox[t]{.4cm}{res5.x\\($\times3$)} & $\ns\begin{matrix} 1^2, 512 \\ 3^2, 512 \\ 1^2, 2048 \end{matrix}$ & \parbox[t]{.6cm}{$7^2$\\$\times2048$} & $\ns\begin{matrix} 3\times1^2, 512 \\ 1\times3^2, 512 \\ 1\times1^2, 2048 \end{matrix}$ & \parbox[t]{1.2cm}{$1\times8^2$\\$\times2048$} & $\ns\begin{matrix} 1^2, 512 \\ 3^2, 512 \\ 1^2, 2048 \end{matrix}$ & \parbox[t]{1.2cm}{$16\times7^2$\\$\times2048$} & $\ns\begin{matrix} 1^2, 512 \\ 3^2, 512 \\ 1^2, 2048 \end{matrix}$ & \parbox[t]{1.5cm}{\frme{$7^2\times2048$}\\\flow{$7^2\times2048$}} & $\ns\begin{matrix} 3\times1^2, \slow{512}|\fast{64} \\ 1\times3^2, \slow{512}|\fast{64} \\ 1\times1^2, \slow{2048}|\fast{256} \end{matrix}$ & \parbox[t]{1.8cm}{\slow{$2\times8^2\times2048$}\\\fast{$16\times8^2\times256$}} \\ [.5cm]
\hline
fusion & & & & & & & & $7^2\times2048$ & & $1\times1^2\times2560$ \\
\hline
\parbox[t]{.6cm}{spatial\\pool} & & $1^2\times2048$ & & $1\times1^2\times2048$ & & $16\times1^2\times2048$ & & $1^2\times2048$ & & \\
\hline
flatten & & $2048$ & & $2048$ & & $16\times2048$ & & $2048$ & & $2560$ \\
\hline
lstm & & & & & & $16\times128$ & & & & \\
\hline
dense & & $2$ & & $2$ & & $16\times 2$ & & $2$ & & $2$ \\
\hline
\end{tabular}
\begin{tablenotes}
\item[a] For 2D CNN and Two-Stream, colors \frme{red} and \flow{blue} highlight how dimensions differ between \frme{frames} and \flow{flows}.
\item[b] For SlowFast, colors $\slow{\text{orange}}|\fast{\text{cyan}}$ highlight the differences in model parameters and dimensions between the \slow{slow} and \fast{fast} pathways.
\item[c] Only for flow input; Serves the purpose of producing 3 channels for transfer learning.
\end{tablenotes}
\end{threeparttable}
\end{table*}

\subsection{Stage I: Models for frame-level probability estimation}
\label{sec:method:sub:models}

In Stage I, our models estimate $p_{intake}$, which is the probability that the label of the target frame is ``intake''.
The four models identified from the literature on video action recognition represent our main models (3D CNN, CNN-LSTM, Two-Stream, SlowFast; see Fig. \ref{fig:architectures}).
In addition to the target frame, each of these four models considers a temporal context of further frames preceding the target frame.
As a baseline and for experiments, we additionally employ a 2D CNN.
Because the 2D CNN does not have a temporal context, this enables us to (1) discern to what extent the temporal context improves model performance and (2) directly compare the importance of appearance and motion features.

For each model, we propose a small instantiation with relatively few parameters, and a larger instantiation using ResNet-50 \cite{he2016deep} as backbone.
In the following, we present each of the proposed models adapted for food intake gesture detection (see Tables \ref{tab:small} and \ref{tab:resnet} for details).
Source code for all models is available at \url{https://github.com/prouast/deep-intake-detection}.

\setcounter{subsubsection}{-1}

\subsubsection{2D CNN}
\label{subsec:2d-cnn}

The 2D CNN functions as a baseline for our study, indicating what is possible without temporal context.
This allows us to discern the importance of the temporal context for intake gesture detection.
Further, the 2D CNN also allows us to directly compare a model based solely on motion to one solely based on visual appearance.
This assessment is not possible for the other four models.

Motion information can be of importance for classes with fast movement such as juggling \cite{karpathy2014large}.
For detection of intake gestures, it seems intuitive that appearance may be the more important modality, which is what we are seeking to confirm here.
For appearance input is \textit{the single target frame}, and for motion \textit{the optical flow between the target frame and the frame directly preceding it}.
We use Dual TV-L$^1$ optical flow \cite{zach2007duality}, which produces two channels of optical flow corresponding to the horizontal and vertical components, as opposed to three RGB channels for frames.

\setlist[description]{font=\normalfont\itshape\space}

\begin{description}
	\item[\arabic{subsubsection}a) Small instantiation] A five-layer CNN of the architecture type popularised by AlexNet \cite{krizhevsky2012imagenet}.
	\item[\arabic{subsubsection}b) ResNet-50 instantiation] We adopt the architecture given by \cite{he2016deep}, which allows us to use pre-trained models.
\end{description}

\subsubsection{3D CNN}
\label{subsec:3d-cnn}

This model has the ability to learn spatio-temporal features.
We extend the 2D CNN introduced in the previous section by using 3D instead of 2D convolutions.
The third dimension corresponds to the temporal context.
We use temporal pooling following the slow fusion approach \cite{karpathy2014large}.

\begin{description}
	\item[\arabic{subsubsection}a) Small instantiation] Extending the small 2D CNN to 3D, we use temporal convolution kernels of size 3 as recommended by \cite{tran2015learning}; temporal pooling is realized in the max pooling layers.
	\item[\arabic{subsubsection}b) ResNet-50 instantiation] We extend ResNet-50 \cite{he2016deep} to 3D, but modify the dimensions to fit our input, since we do not use transfer learning for the 3D CNN. Within each block, the first convolutional layer has a temporal kernel size of 3, a choice adopted from \cite{feichtenhofer2018slowfast}. Temporal fusion is facilitated by using temporal stride 2 in the second convolutional layer of the first block in each block layer.
\end{description}

\subsubsection{CNN-LSTM}
\label{subsec:cnn-lstm}

The CNN-LSTM adds an LSTM layer to model a sequence of high-level features learned from raw frames.
Note that this does not allow the model to learn low-level spatio-temporal features (as opposed to 3D CNN).
Given the clear temporal structure of intake gestures (movement towards the mouth and back), it does however seem intuitive that knowledge of the development of high-level features from temporal context could help predict the current frame.

\begin{description}
	\item[\arabic{subsubsection}a) Small instantiation] We use the features from the first dense layer of the small 2D CNN described previously as input into one LSTM layer with 128 units.
	\item[\arabic{subsubsection}b) ResNet-50 instantiation] The spatially pooled output of a ResNet-50's \cite{he2016deep} last block is used as input into one LSTM layer with 128 units.
\end{description}

\subsubsection{Two-Stream}
\label{subsec:two-stream}

For our instantiations of the Two-Stream approach, we follow the original work by Simonyan and Zisserman \cite{simonyan2014two} to select the model input:
The appearance stream takes the target frame as input;
meanwhile, the motion stream is based on the stacked horizontal and vertical components of optical flow calculated using Dual TV-L$^1$ from pairs of consecutive frames in the temporal context.

\begin{description}
	\item[\arabic{subsubsection}a) Small instantiation] Motion and appearance stream both follow the small 2D CNN architecture; after the last pooling layer, the streams are pooled using spatially aligned conv fusion as proposed by \cite{feichtenhofer2016convolutional}.
	\item[\arabic{subsubsection}b) ResNet-50 instantiation] Motion and appearance stream both follow the ResNet-50 \cite{he2016deep} architecture; after the last block layer, the streams are pooled using spatially aligned conv fusion \cite{feichtenhofer2016convolutional}.
\end{description}

\subsubsection{SlowFast}
\label{subsec:slowfast}

The SlowFast model processes the temporal context at two different temporal resolutions.
Since our dataset has fewer frames than in the original work \cite{feichtenhofer2018slowfast}, we choose the factors $\alpha=4$ and $\beta=0.25$ for our SlowFast instantiations.

\begin{description}
	\item[\arabic{subsubsection}a) Small instantiation] Both pathways are based on the small 2D CNN; we extend the convolutional layers to 3D and set the temporal kernel size to $1$ for the slow pathway and to $3$ for the fast pathway. Following \cite{feichtenhofer2018slowfast}, we choose time-strided convolutions of kernel size $3\times1^2$ for a lateral connection after each of the four convolutional layers. Fusion consists of temporal average pooling and spatially aligned 2D conv fusion \cite{feichtenhofer2016convolutional}.
	\item[\arabic{subsubsection}b) ResNet-50 instantiation] We directly follow \cite{feichtenhofer2018slowfast} who themselves used ResNet-50 as backbone for SlowFast, only using the same dimension tweaks as in our ResNet-50 2D CNN. Fusion consists of global average pooling and concatenation.
\end{description}

\subsection{Loss calculation}
\label{sec:method:sub:loss}

We use cross-entropy loss for all our models.
At evaluation time, we only consider the target frame for prediction, which corresponds to the last frame of the input (see Fig. \ref{fig:stages}).
The same applies to loss calculation during training for all models except CNN-LSTM:
Following \cite{carreira2017quo}, we train the CNN-LSTM using the labels of all input frames, but evaluate only using the label of the target frame.

Due to the nature of our data, the classes are very imbalanced with many more ``non-intake'' frames than ``intake'' frames.
When computing mini-batch loss, we correct for this imbalance by using weights calculated as

\renewcommand{\vec}[1]{\mathbf{#1}}

\begin{equation}
	w_i = \frac{m}{C(i)*n}
\end{equation}

to scale the loss for the $m$ labels $\vec{y}=\{y_1, ..., y_m\}$ in each minibatch, where $n$ is the number of classes and $C(i)$ is the number of elements of $\vec{y}$ which equal $y_i$.

\begin{figure*}[t]
\centering
\includegraphics[width=.8\textwidth]{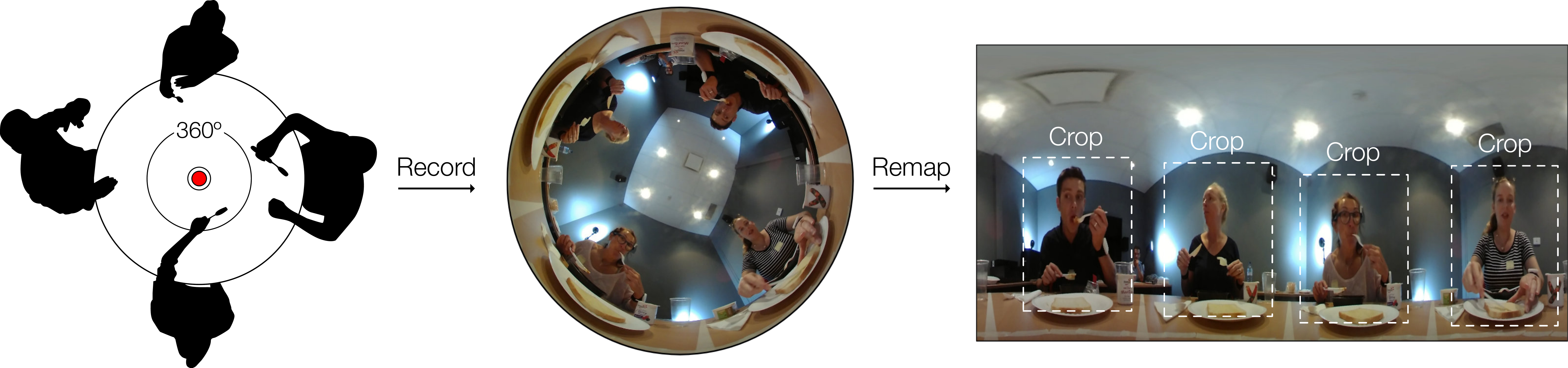}
\caption{Recording of one session. The spherical video is remapped to equirectangular representation, cropped, and reshaped to square shape.}
\label{fig:recording}
\end{figure*}

\begin{table*}[t]
\centering
\caption{Summary statistics for our dataset and the training/validation/test split.}
\label{tab:oreba-summary}
\begin{tabular}{ l | c c c | c c c | c c c | c c c }
& \multicolumn{3}{c|}{Training} & \multicolumn{3}{c|}{Validation} & \multicolumn{3}{c|}{Test} & \multicolumn{3}{c}{Total} \\
Type & \# & Mean [s] & Std [s] & \# & Mean [s] & Std [s] & \# & Mean [s] & Std [s] & \# & Mean [s] & Std [s] \\
\thickhline
Participants & 62 & 802.25 & 243.31 & 20 & 785.97 & 243.74 & 20 & 891.00 & 222.27 & 102 & 816.46 & 240.07 \\
Intake Gestures & 2924 & 2.35 & 1.03 & 952 & 2.28 & 1.00 & 997 & 2.29 & 1.01 & 4891 & 2.32 & 1.02 \\
\hline
\thickhline
\end{tabular}
\end{table*}

\subsection{Stage II: Detecting intake gestures}
\label{sec:method:sub:detect}

We follow a maximum search approach \cite{kyritsis2019modeling} to determine the sparse individual intake gesture times.
Based on estimated frame-level probabilities $\vec{p}$, we first derive $\vec{p'}$ by setting all probabilities below a threshold $p_t$ to zero.
This leaves all frames $\{f:p_{intake,f} \geq p_t\}$ as candidates for detections, as seen at the bottom of Fig. \ref{fig:stages}.
Subsequently, we perform a search for local maxima in $\vec{p'}$ with a minimum distance $d$ between maxima.
The intake gesture times are then inferred from the indices of the maxima.


\section{Dataset}
\label{sec:dataset}

We are not aware of any publicly available dataset including labeled video data of intake gestures.
Related studies that involved collection of video data as ground truth typically do not make the video data available, and instead focus on the inertial \cite{kyritsis2019modeling} and audio sensor data \cite{merck2016multimodality}.

For this research, we collected and labeled video data of 102 participants consuming a standardized meal of lasagna, bread, yogurt, and water in a group setting (ethics approval H-2017-0208).
The data was collected in sessions of four participants at a time seated around a round table in a closed room without outside interference.
Participants were invited to consume their meal in a natural way\footnote{After the meal, 64 of the 102 participants (63\%) responded to the statement ``The presence of the video camera changed my eating behavior'' (5-point Likert scale, ranging from (1) strongly disagree to (5) strongly agree). With an average score of 2.11, we conclude that participants did not feel that the presence of the camera considerably affected their eating behavior.} and encouraged to have a conversation in the process.
A 360fly-4K camera was placed in the center of the table, recording all four participants simultaneously.
As illustrated in Fig. \ref{fig:recording}, raw spherical video was first remapped to equirectangular representation.
We then cropped out a separate video for each individual participant such that the dimensions include typical intake gestures.
Each video was trimmed in time to only include the duration of the meal, and spatially scaled to a square shape.

Two independent annotators labeled and cross-checked the intake gestures in all 102 videos as durations using ChronoViz\footnote{See \url{http://chronoviz.com}.}.
Each gesture is assigned as start timestamp the point where the final uninterrupted movement towards the mouth starts; as end timestamp, it is assigned the point when the participant has finished returning their hand(s) from the movement or started a different action.
Based on the start and end timestamps, we derive a label for each video frame according to the following procedure:
If a video frame was taken between start and end of an annotated gesture, it is assigned the label ``intake''.
If a video frame is taken outside of any annotated gestures, it is assigned the label ``non-intake''.
The dataset is available from the authors on request.


\section{Experiments}
\label{sec:experiments}

\newcommand\tp{\mathit{TP}}
\newcommand\fn{\mathit{FN}}
\newcommand\fp{\mathit{FP}}

\begin{table*}[t]
\centering
\caption{Results for Stages I and II. Reported values are based on the test set, which was only evaluated once.}
\label{tab:results}
\begin{threeparttable}
\begin{tabular}{ l | c c | c | c c | c c c c c c }
Model & \multicolumn{2}{c|}{Features used\tnote{a}} & Temporal & \multicolumn{2}{c|}{Stage I} & \multicolumn{6}{c}{Stage II\tnote{c}} \\
& Frames & Flows & context\tnote{b} & \#Params & UAR & $p\hat{}_t$ & $\tp$ & $\fp_1$ & $\fp_2$ & $\fn$ & $F_1$ \\
\thickhline
0a) \textcolor{small_2d_cnn_frame}{$\blacksquare$} Small 2D CNN & \checkmark & & \multirow{4}{*}{Without} & 4.26M & 82.63\% & 0.957 & 670 & 39 & 287 & 321 & 0.674 \\
\cline{1-3} \cline{5-12}
0a) \textcolor{small_2d_cnn_flow}{$\blacksquare$} Small 2D CNN & & \checkmark & & 4.26M & 71.76\% & 0.793 & 662 & 45 & 1023 & 329 & 0.487 \\
\cline{1-3} \cline{5-12}
0b) \textcolor{resnet_2d_cnn_frame}{$\blacksquare$} ResNet-50 2D CNN & \checkmark & & & 23.5M & 86.39\% & 0.964 & 829 & 54 & 211 & 162 & 0.795 \\
\cline{1-3} \cline{5-12}
0b) \textcolor{resnet_2d_cnn_flow}{$\blacksquare$} ResNet-50 2D CNN & & \checkmark & & 23.5M & 71.34\% & 0.865 & 661 & 53 & 1163 & 330 & 0.461 \\
\thickhline
1a) \textcolor{small_3d_cnn}{$\blacksquare$} Small 3D CNN & \checkmark & & \multirow{8}{*}{With} & 4.39M & 87.54\% & 0.997 & 795 & 37 & 169 & 196 & 0.798 \\
\cline{1-3} \cline{5-12}
2a) \textcolor{small_cnn_lstm}{$\blacksquare$} Small CNN-LSTM & \checkmark & & & 4.85M & 83.36\% & 0.983 & 674 & 17 & 104 & 317 & 0.755 \\
\cline{1-3} \cline{5-12}
3a) \textcolor{small_two_stream}{$\blacksquare$} Small Two-Stream & \checkmark & \checkmark & & 4.34M & 81.96\% & 0.973 & 653 & 36 & 185 & 338 & 0.700 \\
\cline{1-3} \cline{5-12}
4a) \textcolor{small_slowfast}{$\blacksquare$} Small SlowFast & \checkmark & & & 4.49M & 88.71\% & 0.996 & 754 & 31 & 103 & 237 & 0.803 \\
\cline{1-3} \cline{5-12}
1b) \textcolor{resnet_3d_cnn}{$\blacksquare$} ResNet-50 3D CNN & \checkmark & & & 32.2M & 88.77\% & 0.992 & 775 & 25 & 54 & 216 & 0.840 \\
\cline{1-3} \cline{5-12}
2b) \textcolor{resnet_cnn_lstm}{$\blacksquare$} ResNet-50 CNN-LSTM & \checkmark & & & 24.6M & 89.74\% & 0.996 & 791 & 29 & 38 & 200 & 0.856 \\
\cline{1-3} \cline{5-12}
3b) \textcolor{resnet_two_stream}{$\blacksquare$} ResNet-50 Two-Stream\hspace{2mm} & \checkmark & \checkmark & & 47.0M & 85.25\% & 0.997 & 806 & 49 & 82 & 185 & 0.836 \\
\cline{1-3} \cline{5-12}
4b) \textcolor{resnet_slowfast}{$\blacksquare$} ResNet-50 SlowFast & \checkmark & & & 36.7M & 89.01\% & 0.987 & 824 & 23 & 83 & 167 & 0.858 \\
\thickhline
\end{tabular}
\begin{tablenotes}
\item[a] Frame (appearance) features are \textit{raw frames}; Flow (motion) features are \textit{optical flow computed between adjacent frames}.
\item[b] Temporal context consists of 16 frames, the last of which is the target frame.
\item[c] Downsampling to 8 fps causes temporally close events to merge, hence total number of intake gestures in the test set is 991.
\end{tablenotes}
\end{threeparttable}
\end{table*}

We use a global split of our dataset into 62 participants for training, 20 participants for validation, and 20 participants for test as summarized in Table \ref{tab:oreba-summary}.
To reduce computational burden, we downsample the video from 24 fps to 8 fps, and resize to dimensions 140x140 (128x128 after augmentation).

\subsection{Stage I: Estimating frame-level intake probability}
\label{subsec:classification}

We apply the models introduced in Section \ref{sec:method} to classify frames according to the two labels ``intake'' and ``non-intake''.
For our experiments, we distinguish between models without and with temporal context:

\begin{itemize}
	\item Models \textit{without} temporal context (0a-0b) are of interest as a baseline, and to experimentally compare appearance and motion features. For appearance, input is the \textit{single target frame}, and for motion, \textit{optical flow between the target frame and the one preceding it}.
	\item For the models \textit{with} temporal context (1a-4b), input consists of 16 frames, which corresponds to 2 seconds at 8 fps. The last of these frames is the prediction target. To take maximum advantage of the available training data, we generate input using a window shifting by one frame. The use of temporal context implies that the first 15 labels are not predicted.
\end{itemize}

\subsubsection{Training}

We use the Adam optimizer to train each model on the training set.
Training runs for 60 epochs with a learning rate starting at 3e-4 and exponentially decaying at a rate of 0.9 per epoch.
Models without temporal context are trained using batch size 64, while models with temporal context are trained using batch size 8.\footnote{Batch sizes were chosen considering space constraints training on NVIDIA Tesla V100 at fp32 accuracy, and to be consistent across models.}
Using the validation set, we keep track of the best models in terms of unweighted average recall (UAR), which is not biased by class imbalance.
For regularization, we use l2 loss with a lambda of 1e-4.
Dropout is used in all small instantiations of our models on convolutional and dense layers with rate 0.5, but we do not use dropout for the ResNet-50 instantiations.
We also use data augmentation by dynamically applying random transformations: Small rotations, cropping to size 128x128, horizontal flipping, brightness and contrast changes.
All models are learned end-to-end, optical flow is precomputed using Dual TV-L$^1$ \cite{zach2007duality}.

\subsubsection{Transfer learning and warmstarting for better initial parameters}

While the initial small 2D CNN is trained from scratch, we use it to warmstart the convolutional layers of both the small CNN-LSTM and the small Two-Stream model.
The ResNet-50 2D CNN is initialized using an off-the-shelf ResNet-50 trained on the ImageNet database.
To fit ImageNet dimensions, we resize our inputs for this model to 224x224, as listed in Table \ref{tab:resnet}.
We use the ResNet-50 2D CNN to warmstart the convolutional layers of both the ResNet-50 CNN-LSTM and the ResNet-50 Two-Stream model.
All 3D-CNN and SlowFast models are trained from scratch.

\subsection{Stage II: Detecting intake gestures}
\label{subsec:gestures}

For the detection of intake gestures, we build on the exported frame-level probabilities using the models trained in Stage I.
We then apply the approach described in Section \ref{sec:method:sub:detect} to determine sparse detections.

\subsubsection{Evaluation scheme}

\begin{figure}[t]
\centering
\includegraphics[width=.8\columnwidth]{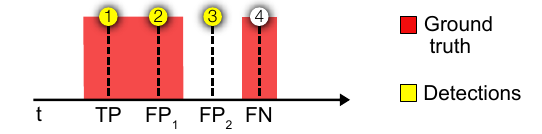}
\caption{The evaluation scheme proposed by Kyritsis et al. \cite{kyritsis2019modeling}. (1) A true positive is the first detection within each ground truth event; (2) False positives of type 1 are further detections within the same ground truth event; (3) False positives of type 2 are detections outside ground truth events; (4) False negatives are non-detected ground truth events.}
\label{fig:scheme}
\end{figure}

We use the evaluation scheme proposed by Kyritsis et al. \cite{kyritsis2019modeling} as seen in Fig. \ref{fig:scheme}.
According to the scheme, one correct detection per ground truth event counts as a true positive ($\tp$), while further detections within the same ground truth event are false positives of type 1 ($\fp_1$).
Detections outside ground truth events are false positives of type 2 ($\fp_2$), and non-detected ground truth events count as false negatives ($\fn$).
Based on the aggregate counts, we calculate precision ($\frac{\tp}{\tp+\fp_1+\fp_2}$), recall ($\frac{\tp}{\tp+\fn}$), and the $F_1$ score ($2*\frac{\mathit{Precision}*\mathit{Recall}}{\mathit{Precision}+\mathit{Recall}}$).

\subsubsection{Parameter setting}

The approach described in Section \ref{sec:method:sub:detect} requires setting two hyperparameters: The minimum distance between detections $d$, and the threshold $p_t$.
We follow Kyritsis et al. \cite{kyritsis2019modeling} and set $d=2s$, which approximates the mean duration of intake gestures, see Table \ref{tab:oreba-summary}.
Since we only run one final evaluation of each model on the test set, we use the validation set to approximate a good threshold $p_t$.
Hence, for each model, we run a grid search between $0.5$ and $1$ on the validation set using a step size of $0.001$ and choose the threshold that maximizes $F_1$.
Table \ref{tab:results} lists the final $\vec{p\hat{}_t}$.

\subsection{Results}

\begin{figure}[t]
	\centering
    \begin{subfigure}{.45\columnwidth}
    	\includegraphics[width=\columnwidth]{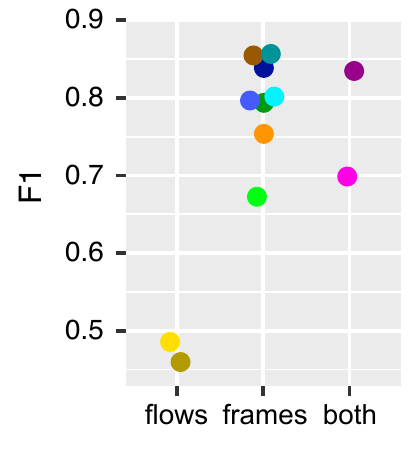}
        \caption{Features used}
        \vspace*{2mm}
    \end{subfigure}%
    \begin{subfigure}{.27\columnwidth}
        \centering
        \includegraphics[width=\columnwidth]{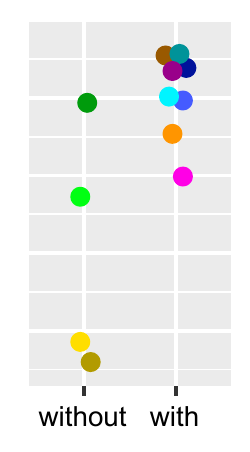}
        \caption{Temporal context}
        \vspace*{2mm}
    \end{subfigure}%
    \begin{subfigure}{.27\columnwidth}
        \centering
        \includegraphics[width=\columnwidth]{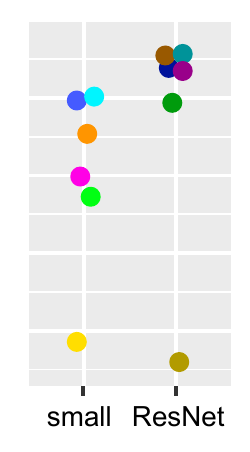}
        \caption{Model depth}
        \vspace*{2mm}
    \end{subfigure}
    \caption{Comparing model performance in terms of $F_1$ scores. It is apparent that (a) models using frames as features perform better than models using optical flow, (b) models with temporal context tend to perform better than models without, and (c) larger (deeper) models tend to perform better. Models are color-coded according to Table \ref{tab:results}.}
    \label{fig:scatter}
\end{figure}

The best result is achieved by the state-of-the-art ResNet-50 SlowFast network with an $F_1$ score of $0.858$.
In general, we find that model accuracy is impacted by three factors of model choice, namely (i) frame or flow features, (ii) with or without temporal context, and (iii) model depth.
Fig. \ref{fig:scatter} illustrates this by plotting the $F_1$ values grouped by each of these factors.

\subsubsection{Frame and flow features, Fig. \ref{fig:scatter} (a)}

Using the 2D CNN, we are able to directly compare how frame (appearance) and flow (motion) features affect model performance.
For the small and ResNet instantiations, frame features lead to a relative $F_1$ improvement of 38\% and 72\% over flow features.
An improvement is also measurable for UAR.
Further, the Two-Stream models, which mainly rely on flow features, perform worse than the other models with temporal context.
We can conclude that for detection of intake gestures, more information is carried by visual appearance than by motion.

\subsubsection{Temporal context, Fig. \ref{fig:scatter} (b)}

To assess the usefulness of temporal context, we compare the accuracies of our models with and without temporal context.
The straightforward extension of Small 2D CNN to Small 3D CNN adds a 17\% relative $F_1$ improvement. 
Comparing the best models with (ResNet-50 SlowFast) and without temporal context (ResNet-50 2D CNN), we find a relative $F_1$ improvement of 8\%. 
We conclude that temporal context considerably improves model accuracy.

Considering model choice, we observe that the Small 3D CNN is superior to its CNN-LSTM counterpart, however the opposite is true for the ResNet-50 instantiations.
This may be due to the fact that for the ResNet instantiations, the CNN-LSTM is pre-trained on ImageNet, while the 3D CNN is not.
We conclude that the 3D CNN could be useful for slim models (e.g., for mobile devices), but for larger models, all architectures with temporal context should be considered.

\subsubsection{Model depth, Fig. \ref{fig:scatter} (c)}

We also see that the deeper ResNet-50 instantiations achieve higher $F_1$ scores than the small ones for all combinations except the flow-based 2D CNN.
Note that the improvement due to model depth is especially noticeable in the $F_1$ score, and less so in UAR.

\subsection{Why do frame features perform better?}

\begin{figure}[t]
\centering
\begin{subfigure}{.384\columnwidth}
	\includegraphics[width=\columnwidth]{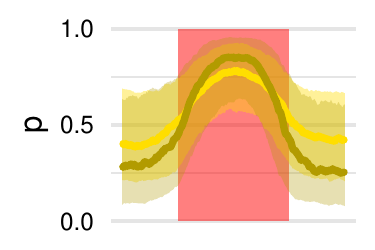}
	\caption{2D CNN (flow)}
\end{subfigure}%
\begin{subfigure}{.308\columnwidth}
	\includegraphics[width=\columnwidth]{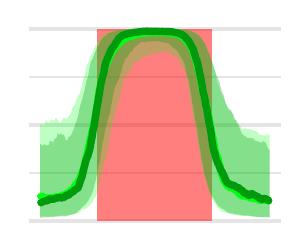}
	\caption{2D CNN (frame)}
\end{subfigure}%
\begin{subfigure}{.308\columnwidth}
	\includegraphics[width=\columnwidth]{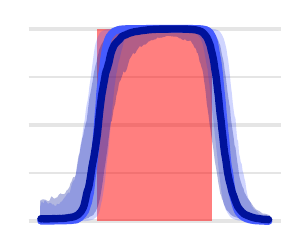}
	\caption{3D CNN}
\end{subfigure}
~
\begin{subfigure}{.384\columnwidth}
	\includegraphics[width=\columnwidth]{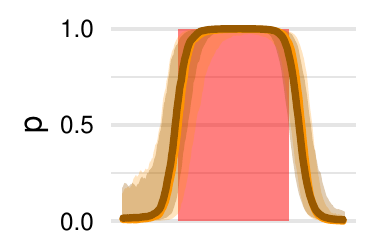}
	\caption{CNN-LSTM}
\end{subfigure}%
\begin{subfigure}{.308\columnwidth}
	\includegraphics[width=\columnwidth]{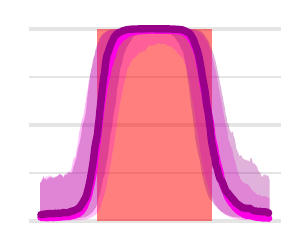}
	\caption{Two-Stream}
\end{subfigure}%
\begin{subfigure}{.308\columnwidth}
	\includegraphics[width=\columnwidth]{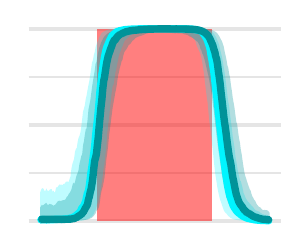}
	\caption{SlowFast}
\end{subfigure}
\caption{Aggregating $p_{intake}$ by model for all ground truth events in the validation set. Predictions have been aligned in time and linearly interpolated. We plot the median and $[q_{25},q_{75}]$ interval for small and ResNet-50 instantiations respectively. Models are color-coded according to Table \ref{tab:results}.}
\label{fig:aligned}
\end{figure}

\begin{figure}[t]
    \centering
    \begin{subfigure}{.5\columnwidth}
        \centering
        \begin{animateinline}[label=myAnim, height=1.8cm, width=4cm, timeline=figures/1010_2/timeline.txt]{8}%
		\hspace{1.8cm}%
		\begin{tikzpicture}
		\node[text width=.2cm] at (.2,0.9) {\scriptsize $p$};
		\draw[color=gray!30] (.4,0) -- (2.2,0) (.4,.45) -- (2.2,.45) (.4,.9) -- (2.2,.9) (.4,1.35) -- (2.2,1.35) (.4,1.8) -- (2.2,1.8);
		\end{tikzpicture}
		\newframe%
		\multiframe{31}{iFrom=0+1,iTo=1+1}{%
		\hspace{2.2cm}%
		\begin{tikzpicture}%
		\begin{axis}[scale only axis, height=1.8cm, width=1.8cm, axis lines*=left, ymin=-0.01, ymax=1.01, xmin=1, xmax=31, hide axis]%
		\addplot[color=small_2d_cnn_flow, mark=none, line width=.4mm, restrict expr to domain={\coordindex}{\iFrom:\iTo}] table[y="prob.oreba_2d_cnn_flow.eval", x="seq", col sep=comma]{figures//1010_2/data.csv};
		\addplot[color=small_2d_cnn_frame, mark=none, line width=.4mm, restrict expr to domain={\coordindex}{\iFrom:\iTo}] table[y="prob.oreba_2d_cnn_frame.eval", x="seq", col sep=comma]{figures//1010_2/data.csv};
		\addplot[color=resnet_2d_cnn_flow, mark=none, line width=.4mm, restrict expr to domain={\coordindex}{\iFrom:\iTo}] table[y="prob.resnet_2d_cnn_flow.eval", x="seq", col sep=comma]{figures//1010_2/data.csv};
		\addplot[color=resnet_2d_cnn_frame, mark=none, line width=.4mm, restrict expr to domain={\coordindex}{\iFrom:\iTo}] table[y="prob.resnet_2d_cnn_frame.eval", x="seq", col sep=comma]{figures//1010_2/data.csv};
		\end{axis}
		\end{tikzpicture}
		}%
		\multiframe{30}{iFrom=1+1}{%
		\includegraphics[width=1.8cm,height=1.8cm]{figures/1010_2/video\iFrom}%
		\hspace{0.4cm}%
		\begin{tikzpicture}
		\draw[-, opacity=0] (0,0)--(1.8,0);%
		\end{tikzpicture}
		}%
		\end{animateinline}
        \caption{Cutting lasagne}
        \vspace*{2mm}
    \end{subfigure}%
    \begin{subfigure}{.5\columnwidth}
        \centering
        \begin{animateinline}[label=myAnim, height=1.8cm, width=4cm, timeline=figures/1015_1/timeline.txt]{8}%
		\hspace{1.8cm}%
		\begin{tikzpicture}
		\node[text width=.2cm] at (.2,0.9) {\scriptsize $p$};
		\draw[color=gray!30] (.4,0) -- (2.2,0) (.4,.45) -- (2.2,.45) (.4,.9) -- (2.2,.9) (.4,1.35) -- (2.2,1.35) (.4,1.8) -- (2.2,1.8);
		\fill[red, opacity=.5] (1.192,0) rectangle (1.912,1.8); 
		\end{tikzpicture}
		\newframe%
		\multiframe{51}{iFrom=0+1,iTo=1+1}{%
		\hspace{2.2cm}%
		\begin{tikzpicture}%
		\begin{axis}[scale only axis, height=1.8cm, width=1.8cm, axis lines*=left, ymin=-0.01, ymax=1.01, xmin=1, xmax=51, hide axis]%
		\addplot[color=small_2d_cnn_flow, mark=none, line width=.4mm, restrict expr to domain={\coordindex}{\iFrom:\iTo}] table[y="prob.oreba_2d_cnn_flow.eval", x="seq", col sep=comma]{figures//1015_1/data.csv};
		\addplot[color=small_2d_cnn_frame, mark=none, line width=.4mm, restrict expr to domain={\coordindex}{\iFrom:\iTo}] table[y="prob.oreba_2d_cnn_frame.eval", x="seq", col sep=comma]{figures//1015_1/data.csv};
		\addplot[color=resnet_2d_cnn_flow, mark=none, line width=.4mm, restrict expr to domain={\coordindex}{\iFrom:\iTo}] table[y="prob.resnet_2d_cnn_flow.eval", x="seq", col sep=comma]{figures//1015_1/data.csv};
		\addplot[color=resnet_2d_cnn_frame, mark=none, line width=.4mm, restrict expr to domain={\coordindex}{\iFrom:\iTo}] table[y="prob.resnet_2d_cnn_frame.eval", x="seq", col sep=comma]{figures//1015_1/data.csv};
		\end{axis}
		\end{tikzpicture}
		}%
		\multiframe{50}{iFrom=1+1}{%
		\includegraphics[width=1.8cm,height=1.8cm]{figures/1015_1/video\iFrom}%
		\hspace{0.4cm}%
		\begin{tikzpicture}
		\draw[-, opacity=0] (0,0)--(1.8,0);%
		\end{tikzpicture}
		}%
		\end{animateinline}
        \caption{Preparing intake}
        \vspace*{2mm}
    \end{subfigure}
    ~
    \begin{subfigure}{\columnwidth}
        \centering
        \includegraphics[width=\columnwidth]{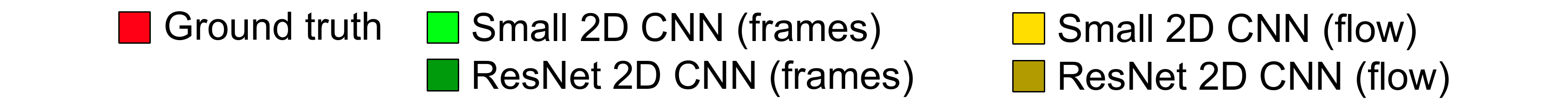}
    \end{subfigure}
    \caption{Example situations showing uncertainty of flow (motion) models compared to frame (appearance) models. Play-on-click in Adobe Acrobat Reader.}
    \label{fig:example-1}
\end{figure}

To help explain why frames perform better as features for this task, we took a closer look at some example model predictions from the validation set.
It appears that flows are in general useful as features; however, the data shows that in comparison to frame models, flow models are less certain about their predictions.
For example, Fig. \ref{fig:example-1} (a) shows how during periods with no intake gestures, small movements such as using cutlery can cause higher uncertainty in flow models.
On the other hand, Fig. \ref{fig:example-1} (b) shows how flow models are also overall less confident when correctly identifying intake gestures.
This can also be observed by looking at aggregated predictions for all events in Fig. \ref{fig:aligned}: Models based solely on flows (a) are less certain about predictions, while their predictions also contain more variance than models based on frames (b).
Further, this is also reflected in the lower thresholds required to trigger a detection for flow models, as is evident from Table \ref{tab:results}.
These lower thresholds and uncertainty are linked to the large number of false positives of these models.

\subsection{Why do models with temporal context perform better?}

\begin{figure}[t]
    \centering
    \begin{subfigure}{.5\columnwidth}
        \centering
        \begin{animateinline}[label=myAnim, height=1.8cm, width=4cm, timeline=figures/1020_2/timeline.txt]{8}%
		\hspace{1.8cm}%
		\begin{tikzpicture}
		\node[text width=.2cm] (p) at (.2,0.9) {\scriptsize $p$};
		\draw[color=gray!30] (.4,0) -- (2.2,0) (.4,.45) -- (2.2,.45) (.4,.9) -- (2.2,.9) (.4,1.35) -- (2.2,1.35) (.4,1.8) -- (2.2,1.8);
		\fill[red, opacity=.5] (0.76,0) rectangle (1.75,1.8); 
		\end{tikzpicture}
		\newframe%
		\multiframe{41}{iFrom=0+1,iTo=1+1}{%
		\hspace{2.2cm}%
		\begin{tikzpicture}%
		\begin{axis}[scale only axis, height=1.8cm, width=1.8cm, axis lines*=left, ymin=-0.01, ymax=1.01, xmin=1, xmax=41, hide axis]%
		\addplot[color=small_2d_cnn_frame, mark=none, line width=.4mm, restrict expr to domain={\coordindex}{\iFrom:\iTo}] table[y="prob.oreba_2d_cnn_frame.eval", x="seq", col sep=comma]{figures/1020_2/data.csv};
		\addplot[color=resnet_2d_cnn_frame, mark=none, line width=.4mm, restrict expr to domain={\coordindex}{\iFrom:\iTo}] table[y="prob.resnet_2d_cnn_frame.eval", x="seq", col sep=comma]{figures/1020_2/data.csv};
		\addplot[color=small_3d_cnn, mark=none, line width=.4mm, restrict expr to domain={\coordindex}{\iFrom:\iTo}] table[y="prob.oreba_3d_cnn.eval", x="seq", col sep=comma]{figures/1020_2/data.csv};
		\addplot[color=resnet_3d_cnn, mark=none, line width=.4mm, restrict expr to domain={\coordindex}{\iFrom:\iTo}] table[y="prob.resnet_3d_cnn.eval", x="seq", col sep=comma]{figures/1020_2/data.csv};
		\end{axis}
		\end{tikzpicture}
		}%
		\multiframe{40}{iFrom=1+1}{%
		\includegraphics[width=1.8cm,height=1.8cm]{figures/1020_2/video\iFrom}%
		\hspace{0.4cm}%
		\begin{tikzpicture}
		\draw[-, opacity=0] (0,0)--(1.8,0);%
		\end{tikzpicture}
		}%
		\end{animateinline}
        \caption{Raised fork}
        \vspace*{2mm}
    \end{subfigure}%
    \begin{subfigure}{.5\columnwidth}
        \centering
        \begin{animateinline}[label=myAnim, height=1.8cm, width=4cm, timeline=figures/1075_1/timeline.txt]{8}%
		\hspace{1.8cm}%
		\begin{tikzpicture}
		\node[text width=.2cm] at (.2,0.9) {\scriptsize $p$};
		\draw[color=gray!30] (.4,0) -- (2.2,0) (.4,.45) -- (2.2,.45) (.4,.9) -- (2.2,.9) (.4,1.35) -- (2.2,1.35) (.4,1.8) -- (2.2,1.8);
		\end{tikzpicture}
		\newframe%
		\multiframe{31}{iFrom=0+1,iTo=1+1}{%
		\hspace{2.2cm}%
		\begin{tikzpicture}%
		\begin{axis}[scale only axis, height=1.8cm, width=1.8cm, axis lines*=left, ymin=-0.01, ymax=1.01, xmin=1, xmax=31, hide axis]%
		\addplot[color=small_2d_cnn_frame, mark=none, line width=.4mm, restrict expr to domain={\coordindex}{\iFrom:\iTo}] table[y="prob.oreba_2d_cnn_frame.eval", x="seq", col sep=comma]{figures/1075_1/data.csv};
		\addplot[color=resnet_2d_cnn_frame, mark=none, line width=.4mm, restrict expr to domain={\coordindex}{\iFrom:\iTo}] table[y="prob.resnet_2d_cnn_frame.eval", x="seq", col sep=comma]{figures/1075_1/data.csv};
		\addplot[color=small_3d_cnn, mark=none, line width=.4mm, restrict expr to domain={\coordindex}{\iFrom:\iTo}] table[y="prob.oreba_3d_cnn.eval", x="seq", col sep=comma]{figures/1075_1/data.csv};
		\addplot[color=resnet_3d_cnn, mark=none, line width=.4mm, restrict expr to domain={\coordindex}{\iFrom:\iTo}] table[y="prob.resnet_3d_cnn.eval", x="seq", col sep=comma]{figures/1075_1/data.csv};
		\end{axis}
		\end{tikzpicture}
		}%
		\multiframe{30}{iFrom=1+1}{%
		\includegraphics[width=1.8cm,height=1.8cm]{figures/1075_1/video\iFrom}%
		\hspace{0.4cm}%
		\begin{tikzpicture}
		\draw[-, opacity=0] (0,0)--(1.8,0);%
		\end{tikzpicture}
		}%
		\end{animateinline}
        \caption{Blowing nose}
        \vspace*{2mm}
    \end{subfigure}
    ~
    \begin{subfigure}{\columnwidth}
        \centering
        \includegraphics[width=\columnwidth]{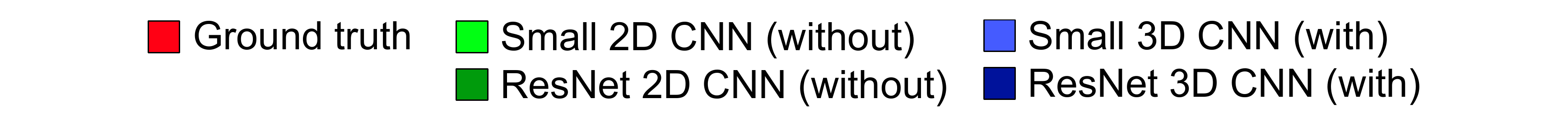}
    \end{subfigure}
    \caption{Example situations where models with temporal context are superior to models without temporal context. Play-on-click in Adobe Acrobat Reader.}
    \label{fig:example-2}
\end{figure}

Our results show that while models based on single frames perform reasonably well, there is measurable improvement when adding temporal context.
Hence, we also looked at this comparison for example model predictions from the validation set to help make the difference easier interpretable.
Indeed, in some cases, it appears intuitive to a human observer how the temporal context is helpful to interpret the target frame.
For example, in Fig. \ref{fig:example-2} (a), the participant keeps the fork raised after completing an intake gesture.
A frame by itself can seem to be part of an intake gesture, while the participant is actually resting this way or is being interrupted.
Without temporal context, the 2D CNN models are unaware of this context, resulting in poor performance.
Availability of temporal context also helps models to become more confident in their predictions.
Further, errors due to outliers are more easily avoidable with temporal context, such as blowing nose in Fig. \ref{fig:example-2} (b).
On the aggregate level, Fig. \ref{fig:aligned} illustrates how predictions by models with temporal context (c)-(f) have a more snug fit with the ground truth events, and less variance in their predictions.

\subsection{Where do the models struggle?}

\begin{figure}[t]
    \centering
    \begin{subfigure}{.5\columnwidth}
        \centering
        \begin{animateinline}[label=myAnim, height=1.8cm, width=4cm, timeline=figures/1004_1/timeline.txt]{8}%
		\hspace{1.8cm}%
		\begin{tikzpicture}
		\node[text width=.2cm] at (.2,0.9) {\scriptsize $p$};
		\draw[color=gray!30] (.4,0) -- (2.2,0) (.4,.45) -- (2.2,.45) (.4,.9) -- (2.2,.9) (.4,1.35) -- (2.2,1.35) (.4,1.8) -- (2.2,1.8);
		\fill [red, opacity=.5] (1.12,0) rectangle (1.78,1.8); 
		\end{tikzpicture}
		\newframe%
		\multiframe{31}{iFrom=0+1,iTo=1+1}{%
		\hspace{2.2cm}%
		\begin{tikzpicture}%
		\begin{axis}[scale only axis, height=1.8cm, width=1.8cm, axis lines*=left, ymin=-0.01, ymax=1.01, xmin=1, xmax=31, hide axis]%
		\addplot[color=resnet_3d_cnn, mark=none, line width=.4mm, restrict expr to domain={\coordindex}{\iFrom:\iTo}] table[y="prob.resnet_3d_cnn.eval", x="seq", col sep=comma]{figures/1004_1/data.csv};
		\addplot[color=resnet_cnn_lstm, mark=none, line width=.4mm, restrict expr to domain={\coordindex}{\iFrom:\iTo}] table[y="prob.resnet_cnn_lstm.eval", x="seq", col sep=comma]{figures/1004_1/data.csv};
		\addplot[color=resnet_slowfast, mark=none, line width=.4mm, restrict expr to domain={\coordindex}{\iFrom:\iTo}] table[y="prob.resnet_slowfast.eval", x="seq", col sep=comma]{figures/1004_1/data.csv};
		\end{axis}
		\end{tikzpicture}
		}%
		\multiframe{30}{iFrom=1+1}{%
		\includegraphics[width=1.8cm,height=1.8cm]{figures/1004_1/video\iFrom}%
		\hspace{0.4cm}%
		\begin{tikzpicture}
		\draw[-, opacity=0] (0,0)--(1.8,0);%
		\end{tikzpicture}
		}%
		\end{animateinline}
        \caption{Eating bread crust}
        \vspace*{2mm}
    \end{subfigure}%
    \begin{subfigure}{.5\columnwidth}
        \centering
        \begin{animateinline}[label=myAnim, height=1.8cm, width=4cm, timeline=figures/1060_2/timeline.txt]{8}%
		\hspace{1.8cm}%
		\begin{tikzpicture}
		\node[text width=.2cm] at (.2,0.9) {\scriptsize $p$};
		\draw[color=gray!30] (.4,0) -- (2.2,0) (.4,.45) -- (2.2,.45) (.4,.9) -- (2.2,.9) (.4,1.35) -- (2.2,1.35) (.4,1.8) -- (2.2,1.8);
		\fill [red, opacity=.5] (0.9625,0) rectangle (1.8625,1.8); 
		\end{tikzpicture}
		\newframe%
		\multiframe{17}{iFrom=0+1,iTo=1+1}{%
		\hspace{2.2cm}%
		\begin{tikzpicture}%
		\begin{axis}[scale only axis, height=1.8cm, width=1.8cm, axis lines*=left, ymin=-0.01, ymax=1.01, xmin=1, xmax=17, hide axis]%
		\addplot[color=resnet_3d_cnn, mark=none, line width=.4mm, restrict expr to domain={\coordindex}{\iFrom:\iTo}] table[y="prob.resnet_3d_cnn.eval", x="seq", col sep=comma]{figures/1060_2/data.csv};
		\addplot[color=resnet_cnn_lstm, mark=none, line width=.4mm, restrict expr to domain={\coordindex}{\iFrom:\iTo}] table[y="prob.resnet_cnn_lstm.eval", x="seq", col sep=comma]{figures/1060_2/data.csv};
		\addplot[color=resnet_slowfast, mark=none, line width=.4mm, restrict expr to domain={\coordindex}{\iFrom:\iTo}] table[y="prob.resnet_slowfast.eval", x="seq", col sep=comma]{figures/1060_2/data.csv};
		\end{axis}
		\end{tikzpicture}
		}%
		\multiframe{16}{iFrom=1+1}{%
		\includegraphics[width=1.8cm,height=1.8cm]{figures/1060_2/video\iFrom}%
		\hspace{0.4cm}%
		\begin{tikzpicture}
		\draw[-, opacity=0] (0,0)--(1.8,0);%
		\end{tikzpicture}
		}%
		\end{animateinline}
        \caption{Licking finger}
        \vspace*{2mm}
    \end{subfigure}
    ~
    \begin{subfigure}{.5\columnwidth}
        \centering
        \begin{animateinline}[label=myAnim, height=1.8cm, width=4cm, timeline=figures/1054_1/timeline.txt]{8}%
		\hspace{1.8cm}%
		\begin{tikzpicture}
		\node[text width=.2cm] at (.2,0.9) {\scriptsize $p$};
		\draw[color=gray!30] (.4,0) -- (2.2,0) (.4,.45) -- (2.2,.45) (.4,.9) -- (2.2,.9) (.4,1.35) -- (2.2,1.35) (.4,1.8) -- (2.2,1.8);
		\fill [red, opacity=.5] (0.4705,0) rectangle (2.1647,1.8); 
		\end{tikzpicture}
		\newframe%
		\multiframe{51}{iFrom=0+1,iTo=1+1}{%
		\hspace{2.2cm}%
		\begin{tikzpicture}%
		\begin{axis}[scale only axis, height=1.8cm, width=1.8cm, axis lines*=left, ymin=-0.01, ymax=1.01, xmin=1, xmax=51, hide axis]%
		\addplot[color=resnet_3d_cnn, mark=none, line width=.4mm, restrict expr to domain={\coordindex}{\iFrom:\iTo}] table[y="prob.resnet_3d_cnn.eval", x="seq", col sep=comma]{figures/1054_1/data.csv};
		\addplot[color=resnet_cnn_lstm, mark=none, line width=.4mm, restrict expr to domain={\coordindex}{\iFrom:\iTo}] table[y="prob.resnet_cnn_lstm.eval", x="seq", col sep=comma]{figures/1054_1/data.csv};
		\addplot[color=resnet_slowfast, mark=none, line width=.4mm, restrict expr to domain={\coordindex}{\iFrom:\iTo}] table[y="prob.resnet_slowfast.eval", x="seq", col sep=comma]{figures/1054_1/data.csv};
		\end{axis}
		\end{tikzpicture}
		}%
		\multiframe{50}{iFrom=1+1}{%
		\includegraphics[width=1.8cm,height=1.8cm]{figures/1054_1/video\iFrom}%
		\hspace{0.4cm}%
		\begin{tikzpicture}
		\draw[-, opacity=0] (0,0)--(1.8,0);%
		\end{tikzpicture}
		}%
		\end{animateinline}
		\caption{Sipping water}
        \vspace*{2mm}
    \end{subfigure}%
    \begin{subfigure}{.5\columnwidth}
        \centering
        \begin{animateinline}[label=myAnim, height=1.8cm, width=4cm, timeline=figures/1112_1/timeline.txt]{8}%
		\hspace{1.8cm}%
		\begin{tikzpicture}
		\node[text width=.2cm] at (.2,0.9) {\scriptsize $p$};
		\draw[color=gray!30] (.4,0) -- (2.2,0) (.4,.45) -- (2.2,.45) (.4,.9) -- (2.2,.9) (.4,1.35) -- (2.2,1.35) (.4,1.8) -- (2.2,1.8);
		\end{tikzpicture}
		\newframe%
		\multiframe{41}{iFrom=0+1,iTo=1+1}{%
		\hspace{2.2cm}%
		\begin{tikzpicture}%
		\begin{axis}[scale only axis, height=1.8cm, width=1.8cm, axis lines*=left, ymin=-0.01, ymax=1.01, xmin=1, xmax=41, hide axis]%
		\addplot[color=resnet_3d_cnn, mark=none, line width=.4mm, restrict expr to domain={\coordindex}{\iFrom:\iTo}] table[y="prob.resnet_3d_cnn.eval", x="seq", col sep=comma]{figures/1112_1/data.csv};
		\addplot[color=resnet_cnn_lstm, mark=none, line width=.4mm, restrict expr to domain={\coordindex}{\iFrom:\iTo}] table[y="prob.resnet_cnn_lstm.eval", x="seq", col sep=comma]{figures/1112_1/data.csv};
		\addplot[color=resnet_slowfast, mark=none, line width=.4mm, restrict expr to domain={\coordindex}{\iFrom:\iTo}] table[y="prob.resnet_slowfast.eval", x="seq", col sep=comma]{figures/1112_1/data.csv};
		\end{axis}
		\end{tikzpicture}
		}%
		\multiframe{40}{iFrom=1+1}{%
		\includegraphics[width=1.8cm,height=1.8cm]{figures/1112_1/video\iFrom}%
		\hspace{0.4cm}%
		\begin{tikzpicture}
		\draw[-, opacity=0] (0,0)--(1.8,0);%
		\end{tikzpicture}
		}%
		\end{animateinline}
        \caption{Too hot}
        \vspace*{2mm}
    \end{subfigure}
    ~
    \begin{subfigure}{\columnwidth}
        \centering
        \includegraphics[width=\columnwidth]{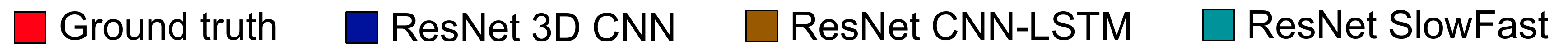}
    \end{subfigure}
    \caption{Example situations where the best models cause false negatives (a-b) and false positives of type 1 (c) and type 2 (d). Note that a true positive always preceeds false positives of type 1. Play-on-click in Adobe Acrobat Reader.}
    \label{fig:example-3}
\end{figure}

Examining the results in Table \ref{tab:results}, it appears that mainly false negatives and also false positives of type 2 are problematic for our best models.
To help understand in what circumstances the models struggle, we compiled examples from the validation set where the best models make mistakes.
The examples show that these mistakes tend to happen in cases of ``outlier behavior'' that differs substantially from the typical behavior in the dataset.
False negatives occur mostly for intake gestures that are less noticeable or less common, such as eating bread crust or licking a finger, see Fig. \ref{fig:example-3} (a) and (b).
An example for false positives of type 2 is when the participant interrupts an intake gesture as depicted in Fig. \ref{fig:example-3} (d).
We see false positives of type 1 as mostly representing a shortcoming of the Stage II approach, i.e., when the duration of an intake gesture exceeds 2 seconds, seen in Fig. \ref{fig:example-3} (c).


\section{Conclusion}
\label{sec:conclusion}

In this paper, we have demonstrated the feasibility of detecting intake gestures from video sourced through a 360-degree camera.
Our two-stage approach involves learning frame-level probabilities using deep architectures proposed in the context of video action recognition (Stage I), and a search algorithm to detect individual intake gestures (Stage II).
Through evaluation of a variety of models, our results show that appearance features in form of the individual raw frames are well suited for this task.
Further, while single frames on their own can lead to useful results with $F_1$ of up to 0.795, the best model considering a temporal context of multiple frames achieves a superior $F_1$ of 0.858.
This result is achieved with a state-of-the-art SlowFast network \cite{feichtenhofer2018slowfast} using ResNet-50 as backbone.

Overall, we see several benefits and opportunities that the use of video holds for dietary monitoring.
First, the proliferation of 360 degree video reduces the practical challenges of recording images of human eating occasions.
This could be used to capture the intake of multiple individuals with a single camera positioned in the center of a table (e.g., families eating from a shared dish \cite{burrows2019dietary}).
Second, the models could be leveraged to support dietitians in reviewing videos of intake occasions.
For instance, instead of watching a twenty minute video, imagery of the actual intake gestures could be automatically extracted for assessment.
Third, the models could be used to semi-automate the ground truth annotation process (e.g., for inertial sensors) by pre-annotating the videos.
Finally, the models could be used to further the development of fully automated dietary monitoring \cite{hantke2016hear} (e.g., care-taking robots, life-logging, patient monitoring).

As a limitation of our approach, we noted that the distribution of participant behavior has a ``fat tail'' as it includes many examples of outlier behavior that models misinterpret (e.g., sudden interruption due to a conversation, blowing on food).
To deal with such events, future research may employ larger databases of samples to train models.
Further, in comparison to approaches based on inertial sensors, our approach has a limitation in that it requires the participant to consume their meal at a table equipped with a camera.
Hence, our vision models should be directly benchmarked against models based on inertial sensor data to determine their relative strengths and weaknesses.
Going one step further, fusion of both modalities could also be explored.
Finally, Stages I and II could be unified into a single end-to-end learning model using CTC loss \cite{graves2006connectionist}, which may alleviate some of the shortcomings of the current approach.
However, it needs to be considered that (i) this is directly only feasible for the CNN-LSTM model without increasing the requirement for GPU memory, and (ii) a larger temporal context and dataset may be required.

\section*{Acknowledgment}

We gratefully acknowledge the support by the Bill \& Melinda Gates Foundation [OPP1171389].
This work was additionally supported by an Australian Government Research Training (RTP) Scholarship.

\bibliographystyle{assets/IEEEtran}
\bibliography{paper}

\begin{thebibliography}{10}
\providecommand{\url}[1]{#1}
\csname url@samestyle\endcsname
\providecommand{\newblock}{\relax}
\providecommand{\bibinfo}[2]{#2}
\providecommand{\BIBentrySTDinterwordspacing}{\spaceskip=0pt\relax}
\providecommand{\BIBentryALTinterwordstretchfactor}{4}
\providecommand{\BIBentryALTinterwordspacing}{\spaceskip=\fontdimen2\font plus
\BIBentryALTinterwordstretchfactor\fontdimen3\font minus
  \fontdimen4\font\relax}
\providecommand{\BIBforeignlanguage}[2]{{%
\expandafter\ifx\csname l@#1\endcsname\relax
\typeout{** WARNING: IEEEtran.bst: No hyphenation pattern has been}%
\typeout{** loaded for the language `#1'. Using the pattern for}%
\typeout{** the default language instead.}%
\else
\language=\csname l@#1\endcsname
\fi
#2}}
\providecommand{\BIBdecl}{\relax}
\BIBdecl

\bibitem{weekes2009review}
C.~Weekes, A.~Spiro, C.~Baldwin, K.~Whelan, J.~Thomas, D.~Parkin, and P.~Emery,
  ``A review of the evidence for the impact of improving nutritional care on
  nutritional and clinical outcomes and cost,'' \emph{J. Human Nutrition
  Dietetics}, vol.~22, pp. 324--335, 2009.

\bibitem{rouast2018using}
P.~V. Rouast, M.~T.~P. Adam, T.~Burrows, R.~Chiong, and M.~E. Rollo, ``Using
  deep learning and 360 video to detect eating behavior for user assistance
  systems,'' in \emph{Proc. Europ. Conf. Information Systems}, 2018.

\bibitem{who2017non}
WHO, ``Noncommunicable diseases progress monitor, 2017,'' Geneva: World Health
  Organization, Tech. Rep., 2017.

\bibitem{lichtman1992discrepancy}
S.~W. Lichtman, K.~Pisarska, E.~R. Berman, M.~Pestone, H.~Dowling,
  E.~Offenbacher, H.~Weisel, S.~Heshka, D.~E. Matthews, and S.~B. Heymsfield,
  ``Discrepancy between self-reported and actual caloric intake and exercise in
  obese subjects,'' \emph{New England J. Medicine}, vol. 327, no.~27, pp.
  1893--1898, 1992.

\bibitem{vu2017wearable}
T.~Vu, F.~Lin, N.~Alshurafa, and W.~Xu, ``Wearable food intake monitoring
  technologies: A comprehensive review,'' \emph{Computers}, vol.~6, no.~1,
  p.~4, 2017.

\bibitem{kyritsis2019modeling}
K.~Kyritsis, C.~Diou, and A.~Delopoulos, ``Modeling wrist micromovements to
  measure in-meal eating behavior from inertial sensor data,'' \emph{IEEE J.
  Biomedical and Health Informatics}, 2019.

\bibitem{hantke2016hear}
S.~Hantke, F.~Weninger, R.~Kurle, F.~Ringeval, A.~Batliner, A.~E.-D. Mousa, and
  B.~Schuller, ``I hear you eat and speak: Automatic recognition of eating
  condition and food type, use-cases, and impact on asr performance,''
  \emph{PloS one}, vol.~11, no.~5, p. e0154486, 2016.

\bibitem{thomaz2015practical}
E.~Thomaz, I.~Essa, and G.~D. Abowd, ``A practical approach for recognizing
  eating moments with wrist-mounted inertial sensing,'' in \emph{Proc.
  UbiComp}.\hskip 1em plus 0.5em minus 0.4em\relax ACM, 2015, pp. 1029--1040.

\bibitem{mirtchouk2016automated}
M.~Mirtchouk, C.~Merck, and S.~Kleinberg, ``Automated estimation of food type
  and amount consumed from body-worn audio and motion sensors,'' in \emph{Proc.
  UbiComp}, 2016, pp. 451--462.

\bibitem{braeken2016secure}
A.~Braeken, P.~Porambage, A.~Gurtov, and M.~Ylianttila, ``Secure and efficient
  reactive video surveillance for patient monitoring,'' \emph{Sensors},
  vol.~16, no.~1, pp. 1--13, 2016.

\bibitem{hall2017implementing}
A.~Hall, C.~B. Wilson, E.~Stanmore, and C.~Todd, ``Implementing monitoring
  technologies in care homes for people with dementia: a qualitative
  exploration using normalization process theory,'' \emph{Int. J. Nursing
  Studies}, vol.~72, pp. 60--70, 2017.

\bibitem{lecun2015deep}
Y.~LeCun, Y.~Bengio, and G.~Hinton, ``Deep learning,'' \emph{Nature}, vol. 521,
  pp. 436--444, 2015.

\bibitem{ji20133d}
S.~Ji, W.~Xu, M.~Yang, and K.~Yu, ``3d convolutional neural networks for human
  action recognition,'' \emph{IEEE Trans. Pattern Anal. Mach. Intell.},
  vol.~35, no.~1, pp. 221--231, 2013.

\bibitem{donahue2015long}
J.~Donahue, L.~Anne~Hendricks, S.~Guadarrama, M.~Rohrbach, S.~Venugopalan,
  K.~Saenko, and T.~Darrell, ``Long-term recurrent convolutional networks for
  visual recognition and description,'' in \emph{Proc. CVPR}, 2015, pp.
  2625--2634.

\bibitem{simonyan2014two}
K.~Simonyan and A.~Zisserman, ``Two-stream convolutional networks for action
  recognition in videos,'' in \emph{Proc. NIPS}, 2014, pp. 568--576.

\bibitem{feichtenhofer2018slowfast}
C.~Feichtenhofer, H.~Fan, J.~Malik, and K.~He, ``Slowfast networks for video
  recognition,'' \emph{arXiv preprint arXiv:1812.03982}, 2018.

\bibitem{ciocca2017learning}
G.~Ciocca, P.~Napoletano, and R.~Schettini, ``Learning cnn-based features for
  retrieval of food images,'' in \emph{Proc. Int. Conf. Image Analysis and
  Processing}, 2017, pp. 426--434.

\bibitem{feichtenhofer2018have}
C.~Feichtenhofer, A.~Pinz, R.~P. Wildes, and A.~Zisserman, ``What have we
  learned from deep representations for action recognition?'' in \emph{Proc.
  CVPR}, 2018, pp. 7844--7853.

\bibitem{karpathy2014large}
A.~Karpathy, G.~Toderici, S.~Shetty, T.~Leung, R.~Sukthankar, and L.~Fei-Fei,
  ``Large-scale video classification with convolutional neural networks,'' in
  \emph{Proc. CVPR}, 2014, pp. 1725--1732.

\bibitem{carreira2017quo}
J.~Carreira and A.~Zisserman, ``Quo vadis, action recognition? a new model and
  the kinetics dataset,'' in \emph{Proc. CVPR}, 2017, pp. 4299--4308.

\bibitem{block1982review}
G.~Block, ``A review of validations of dietary assessment methods,'' \emph{Am.
  J. Epidemiology}, vol. 115, no.~4, pp. 492--505, 1982.

\bibitem{chen2016deep}
J.~Chen and C.-W. Ngo, ``Deep-based ingredient recognition for cooking recipe
  retrieval,'' in \emph{Proc. Multimedia Conf.}, 2016, pp. 32--41.

\bibitem{ciocca2018cnn}
G.~Ciocca, P.~Napoletano, and R.~Schettini, ``Cnn-based features for retrieval
  and classification of food images,'' \emph{Comput. Vision Image
  Understanding}, vol. 176, pp. 70--77, 2018.

\bibitem{puri2009recognition}
M.~Puri, Z.~Zhu, Q.~Yu, A.~Divakaran, and H.~Sawhney, ``Recognition and volume
  estimation of food intake using a mobile device,'' in \emph{Proc. Workshop on
  Applications of Computer Vision}, 2009, pp. 1--8.

\bibitem{zhang2015snap}
W.~Zhang, Q.~Yu, B.~Siddiquie, A.~Divakaran, and H.~Sawhney, ````snap-n-eat''
  food recognition and nutrition estimation on a smartphone,'' \emph{J.
  Diabetes Science Technol.}, vol.~9, no.~3, pp. 525--533, 2015.

\bibitem{dong2014detecting}
Y.~Dong, J.~Scisco, M.~Wilson, E.~Muth, and A.~Hoover, ``Detecting periods of
  eating during free-living by tracking wrist motion,'' \emph{IEEE J.
  Biomedical and Health Informatics}, vol.~18, no.~4, pp. 1253--1260, 2014.

\bibitem{ye2016assisting}
X.~Ye, G.~Chen, Y.~Gao, H.~Wang, and Y.~Cao, ``Assisting food journaling with
  automatic eating detection,'' in \emph{Proc. CHI Conf. Extended Abstracts on
  Human Factors in Computing Systems}.\hskip 1em plus 0.5em minus 0.4em\relax
  ACM, 2016, pp. 3255--3262.

\bibitem{robinson2014systematic}
E.~Robinson, E.~Almiron-Roig, F.~Rutters, C.~de~Graaf, C.~G. Forde,
  C.~Tudur~Smith, S.~J. Nolan, and S.~A. Jebb, ``A systematic review and
  meta-analysis examining the effect of eating rate on energy intake and
  hunger,'' \emph{Am. J. Clinical Nutrition}, vol. 100, no.~1, pp. 123--151,
  2014.

\bibitem{heydarian2019assessing}
H.~Heydarian, M.~Adam, T.~Burrows, C.~Collins, and M.~E. Rollo, ``Assessing
  eating behaviour using upper limb mounted motion sensors: A systematic
  review,'' \emph{Nutrients}, vol.~11, no.~5, p. 1168, 2019.

\bibitem{amft2005analysis}
O.~Amft, M.~Stager, P.~Lukowicz, and G.~Troster, ``Analysis of chewing sounds
  for dietary monitoring,'' in \emph{Proc. UbiComp}, 2005, pp. 56--72.

\bibitem{passler2012food}
S.~P{\"a}{\ss}ler, M.~Wolff, and W.-J. Fischer, ``Food intake monitoring: an
  acoustical approach to automated food intake activity detection and
  classification of consumed food,'' \emph{Physiological Measurement}, vol.~33,
  no.~6, pp. 1073--1093, 2012.

\bibitem{sazonov2012sensor}
E.~S. Sazonov and J.~M. Fontana, ``A sensor system for automatic detection of
  food intake through non-invasive monitoring of chewing,'' \emph{IEEE Sensors
  Journal}, vol.~12, no.~5, pp. 1340--1348, 2012.

\bibitem{amft2005detection}
O.~Amft, H.~Junker, and G.~Troster, ``Detection of eating and drinking arm
  gestures using inertial body-worn sensors,'' in \emph{Proc. Int. Symp.
  Wearable Computers}.\hskip 1em plus 0.5em minus 0.4em\relax IEEE, 2005, pp.
  160--163.

\bibitem{shen2017assessing}
Y.~Shen, J.~Salley, E.~Muth, and A.~Hoover, ``Assessing the accuracy of a wrist
  motion tracking method for counting bites across demographic and food
  variables,'' \emph{IEEE J. Biomedical and Health Informatics}, vol.~21,
  no.~3, pp. 599--606, 2017.

\bibitem{zhang2018sense}
S.~Zhang, W.~Stogin, and N.~Alshurafa, ``I sense overeating: Motif-based
  machine learning framework to detect overeating using wrist-worn sensing,''
  \emph{Information Fusion}, vol.~41, pp. 37--47, 2018.

\bibitem{gao2004dining}
J.~Gao, A.~G. Hauptmann, A.~Bharucha, and H.~D. Wactlar, ``Dining activity
  analysis using a hidden markov model,'' in \emph{Proc. Int. Conf. Pattern
  Recognition}.\hskip 1em plus 0.5em minus 0.4em\relax IEEE, 2004, pp.
  915--918.

\bibitem{okamoto2016grillcam}
K.~Okamoto and K.~Yanai, ``Grillcam: A real-time eating action recognition
  system,'' in \emph{Proc. Int. Conf. Multimedia Modeling}.\hskip 1em plus
  0.5em minus 0.4em\relax Springer, 2016, pp. 331--335.

\bibitem{hondori2012monitoring}
H.~M. Hondori, M.~Khademi, and C.~V. Lopes, ``Monitoring intake gestures using
  sensor fusion (microsoft kinect and inertial sensors) for smart home
  tele-rehab setting,'' in \emph{Proc. Healthcare Innovation Conf.}, 2012, pp.
  36--39.

\bibitem{tham2014automatic}
J.~S. Tham, Y.~C. Chang, and M.~F.~A. Fauzi, ``Automatic identification of
  drinking activities at home using depth data from rgb-d camera,'' in
  \emph{Proc. Int. Conf. Control, Automation and Information Sciences}.\hskip
  1em plus 0.5em minus 0.4em\relax IEEE, 2014, pp. 153--158.

\bibitem{klaser2008spatio}
A.~Klaser, M.~Marsza{\l}ek, and C.~Schmid, ``A spatio-temporal descriptor based
  on 3d-gradients,'' in \emph{Proc. BMVC}, 2008, pp. 1--10.

\bibitem{wang2011action}
H.~Wang, A.~Kl{\"a}ser, C.~Schmid, and L.~Cheng-Lin, ``Action recognition by
  dense trajectories,'' in \emph{Proc. CVPR}, 2011, pp. 3169--3176.

\bibitem{tran2015learning}
D.~Tran, L.~Bourdev, R.~Fergus, L.~Torresani, and M.~Paluri, ``Learning
  spatiotemporal features with 3d convolutional networks,'' in \emph{Proc.
  ICCV}, 2015, pp. 4489--4497.

\bibitem{hochreiter1997long}
S.~Hochreiter and J.~Schmidhuber, ``Long short-term memory,'' \emph{Neural
  Comput.}, vol.~9, no.~8, pp. 1735--1780, 1997.

\bibitem{feichtenhofer2016convolutional}
C.~Feichtenhofer, A.~Pinz, and A.~Zisserman, ``Convolutional two-stream network
  fusion for video action recognition,'' in \emph{Proc. CVPR}, 2016, pp.
  1933--1941.

\bibitem{he2016deep}
K.~He, X.~Zhang, S.~Ren, and J.~Sun, ``Deep residual learning for image
  recognition,'' in \emph{Proc. CVPR}, 2016, pp. 770--778.

\bibitem{zach2007duality}
C.~Zach, T.~Pock, and H.~Bischof, ``A duality based approach for realtime tv-l1
  optical flow,'' in \emph{DAGM: Pattern Recognition}, 2007, pp. 214--223.

\bibitem{krizhevsky2012imagenet}
A.~Krizhevsky, I.~Sutskever, and G.~E. Hinton, ``Imagenet classification with
  deep convolutional neural networks,'' in \emph{Proc. NIPS}, 2012, pp.
  1097--1105.

\bibitem{merck2016multimodality}
C.~Merck, C.~Maher, M.~Mirtchouk, M.~Zheng, Y.~Huang, and S.~Kleinberg,
  ``Multimodality sensing for eating recognition,'' in \emph{Proc. Int. Conf.
  Pervasive Computing Technologies for Healthcare}, 2016, pp. 130--137.

\bibitem{burrows2019dietary}
T.~Burrows, C.~Collins, M.~T.~P. Adam, K.~Duncanson, and M.~Rollo, ``Dietary
  assessment of shared plate eating: A missing link,'' \emph{Nutrients},
  vol.~11, no.~4, pp. 1--14, 2019.

\bibitem{graves2006connectionist}
A.~Graves, S.~Fernandez, F.~Gomez, and J.~Schmidhuber, ``Connectionist temporal
  classification: labelling unsegmented sequence data with recurrent neural
  networks,'' in \emph{Proc. ICML}, 2006, pp. 369--376.

\end{thebibliography}

\vskip -2\baselineskip plus -1fil

\begin{IEEEbiography}[{\includegraphics[width=1in,height=1.25in,clip,keepaspectratio]{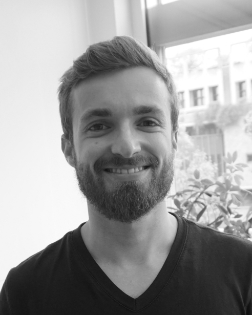}}]{Philipp V. Rouast}
received the B.Sc. and M.Sc. degrees in Industrial Engineering from Karlsruhe Institute of Technology, Germany, in 2013 and 2016 respectively.
He is currently working towards the Ph.D. degree in Information Systems and is a graduate research assistant at The University of Newcastle, Australia.
His research interests include deep learning, affective computing, HCI, and related applications of computer vision.
Find him at \url{https://www.rouast.com}.
\end{IEEEbiography}

\vskip -2\baselineskip plus -1fil

\begin{IEEEbiography}[{\includegraphics[width=1in,height=1.25in,clip,keepaspectratio]{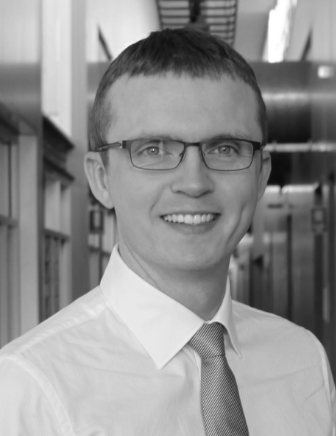}}]{Marc T. P. Adam}
is a Senior Lecturer in Computing and Information Technology at the University of Newcastle, Australia.
In his research, he investigates the interplay of human users' cognition and affect in human-computer interaction.
He is a founding member of the Society for NeuroIS.
He received an undergraduate degree in Computer Science from the University of Applied Sciences W{\"u}rzburg, Germany, and a PhD in Economics of Information Systems from Karlsruhe Institute of Technology, Germany.
\end{IEEEbiography}

\end{document}